\documentclass[eng]{class}
\usepackage{amsmath, amssymb, amsthm}
\usepackage{graphicx}
\usepackage{pdflscape}
\usepackage{afterpage}
\usepackage{blkarray}
\usepackage{bm}
\usepackage{comment}
\usepackage{algorithm}
\usepackage{algpseudocode}
\usepackage{setspace}
\usepackage{xcolor}
\usepackage{atbegshi}
\AtBeginDocument{\AtBeginShipoutNext{\AtBeginShipoutDiscard}}\addtocounter{page}{-1}
\usepackage[backend = biber, style = authoryear, maxbibnames = 99, maxcitenames = 99]{biblatex}
\DeclareFieldFormat[article]{volume}{,\bibstring{volume}\addspace #1,}
\DeclareFieldFormat[article]{number}{\bibstring{number}\addspace #1}
\DefineBibliographyStrings{english}{
    volume = { vol\adddot},
    number = { no\adddot}
    }
    \DeclareFieldFormat{bibentrysetcount}{\mkbibparens{\color{black}#1\addthinspace\bibstring{references}}}
\usepackage{csquotes}
\usepackage{url}
\usepackage{bbm}
\usepackage{verbatim}
\usepackage{dsfont}

\title{Mixed-type Distance Shrinkage and Selection for Clustering via Kernel Metric Learning}

\author[1]{Jesse S. Ghashti}
\author[1]{John R. J. Thompson}

\affil[1]{University of British Columbia, Department of Computer Science, Mathematics, Physics, and Statistics, Kelowna, BC, Canada}

\firstauthor{Ghashti and Thompson}

\contactauthor{John Thompson, 3187 University Way, Kelowna, British Columbia, Canada, V1V 1V7}
\email{john.thompson@ubc.ca}                
\begin{document}

\maketitle
\printcontactdata

\section*{Abstract}

\noindent Distance-based clustering and classification are widely used in various fields to group mixed numeric and categorical data. \textcolor{black}{In many algorithms}, a predefined distance measurement is used to cluster data points based \textcolor{black}{on} their dissimilarity. While there exist numerous distance-based measures for data with pure numerical attributes and several ordered and unordered categorical metrics, an \textcolor{black}{efficient and accurate} distance for mixed-type data \textcolor{black}{that utilizes the continuous and discrete properties simulatenously} is an open problem. Many metrics convert numerical attributes to categorical ones or vice versa. They handle the data points as a single attribute type or calculate a distance between each attribute separately and add them up. We propose a metric \textcolor{black}{called KDSUM} that uses mixed kernels to measure dissimilarity, with cross-validated optimal bandwidth \textcolor{black}{selection}. \textcolor{black}{We demonstrate that KDSUM is a shrinkage method from existing mixed-type metrics to a uniform dissimilarity metric, and} improves clustering accuracy when utilized \textcolor{black}{in} existing distance-based clustering algorithms on simulated and real-world datasets containing continuous\textcolor{black}{-only}, categorical\textcolor{black}{-only}, and mixed-type data. \newline

\noindent \underline{Keywords}: Mixed-type data, metric learning, clustering, smoothing, kernel \textcolor{black}{methods}, similarity measure\textcolor{black}{s} 

\newpage

\section{Introduction}
Datasets comprising continuous, ordered, and unordered categorical data are known as mixed-type data and are prevalent across various disciplines, and the availability of such heterogeneous data types continues to increase. Although several approaches have been employed to calculate the distance for mixed-type data points, there is no broadly accepted definition \textcolor{black}{of distance}. The challenge of quantifying distance is balancing the contributions of each variable--particularly between discrete and continuous--to the overall difference between data entries.  In this paper, we develop a data-driven distance method that estimates the importance of discrete and continuous variables to the difference between entries \textcolor{black}{using a shrinkage approach}. \\

Many existing distances homogenize mixed-type data to single-type by projecting all data to either discrete or continuous, through methods such as discretization or dummy coding before calculating distance (see, for example, Guha et al., 2000; Dougherty et al., 1995). While these distances are computationally efficient and well-known, they can inaccurately calculate the meaningful differences between data points and overweight variables in \textcolor{black}{the} continuous or discrete domains. This overweighting can severely affect the \textcolor{black}{clustering outcome} of any methodology that requires distances through a significant loss of information on the homogenized data types. \\

Clustering is a fundamental technique in data analysis that involves grouping similar data points together based on distance. When clustering mixed-type data, choosing an appropriate distance metric that can handle the heterogeneity of the data types and scales is crucial. The metric should \textcolor{black}{utilize and balance} each data type \textcolor{black}{appropriately to provide a meaningful} distance between data points \textcolor{black}{that accurately represents the similarity and dissimilarity between data points within a particular dataset}. The choice of metric can have a significant impact on the accuracy, reliability, and interpretability of \textcolor{black}{distance-based} clustering results\textcolor{black}{, and it} is essential to carefully consider its performance using the standard  \textcolor{black}{statistics} of clustering accuracy (CA) and Adjusted Rand Index (ARI) (Hubert \& Arabie, 1985). \\


\textcolor{black}{W}e propose a novel kernel distance for mixed-type data that balances mixed-type data \textcolor{black}{important to within-dataset similarity for} clustering applications. \textcolor{black}{We prove that kernel similarity functions can be used as a distance metric and our specific kernel distance metric (called KDSUM) is a shrinkage method between maximized dissimilarity and uniform similarity between all points.} We \textcolor{black}{demonstrate that maximum similarity cross-validation chooses optimal bandwidths}. The advantage of this method is that the importance of each variable to \textcolor{black}{similarity} between \textcolor{black}{data points are balanced by the magnitudes of} cross-validat\textcolor{black}{ed bandwidths}. \\

\textcolor{black}{We apply kernel distance to agglomerative hierarchical clustering and demonstrate the utility of our metric for clustering both simulated and real-world datasets.} We find a kernel distance almost unilaterally improves clustering performance compared to other common mixed-type distances, such as Gower's distance \textcolor{black}{(Gower 1971)}. A kernel distance provides researchers and practitioners with a unified, robust, effective, and efficient distance for mixed-type data, aiding in informed decision-making from the more accurately characterized clusters. \textcolor{black}{We compare KDSUM with agglomerative cluster to competing clustering approaches. For continuous data, we compare to hierarchical clustering techniques with standard linkage methods and Partitioning Around Medoids with Euclidean distance, $k-$means, and Gaussian Mixture models. For categorical data, we compare to hierarchical clustering techniques with standard linkage methods and Partitioning Around Medoids with Gower's distance, $k-$modes, and Robust Clustering using links (ROCK). For mixed-type data, we compare to hierarchical clustering techniques with standard linkage methods and Partitioning Around Medoids with Gower's distance, $k-$prototypes, and model based clustering for mixed-type data (clustMD).}  \\

The paper is structured as follows: Section \ref{sec:relatedWorks} discusses existing homogenized and non-homogenized approaches for mixed-type distances. \textcolor{black}{Section \ref{sec:kernel} describes the kernel methods derived from probability density estimation used in this paper}. Section \ref{sec:methods} contains the methodology for the \textcolor{black}{kernel distance metric (KDSUM)}. \textcolor{black}{Sections \ref{sec:data} and \ref{sec:real} describe the clustering algorithms, evaluation metrics, and the simulated and real data}. Finally, Section \ref{sec:conclusions} \textcolor{black}{are the} conclusion\textcolor{black}{s} and insights for future work.\\

\section{\textcolor{black}{Mixed-type distances}} \label{sec:relatedWorks}
Consider a $n \times p$ mixed-type data matrix $X$ consisting of $n$ observations with $p$ many variables that are a combination of continuous, unordered and ordered categorical variables. Assume that the $p$ variables are arranged such that the first continuous variables $p_c$ are first, followed by unordered categorical variables $p_u$, and then the ordered categorical variables $p_o$ such that $p = p_c+p_u+p_o$. \textcolor{black}{For mixed\textcolor{black}{-type} distances, consider that the rows of $X$ are observation vectors $\mathbf{x}_j$, and the dissimilarity or distance between any two observations $\mathbf{x}_i$ and $\mathbf{x}_{j}$ is denoted $d(\mathbf{x}_i,\mathbf{x}_{j})$, whereas the similarity between the observations is denoted $s(\mathbf{x}_i,\mathbf{x}_{j})$.}\\

Typical methodologies for calculating mixed-type distance require \textcolor{black}{data to be} homogenize\textcolor{black}{d to} either numerical or categorical type. Discretization is the process of converting continuous variables into discrete categories or intervals\textcolor{black}{; an example of this is } binning \textcolor{black}{ which } divid\textcolor{black}{es} the range of a continuous variable into intervals and assign\textcolor{black}{s} each observation to the corresponding interval (Dougherty et al., 1995). For example, a person's age can be binned into discrete ordered categories such as "0-18", "19-30", "31-50", and so on.  \\

Let $\mathbf{x}_i$ be \textcolor{black}{an} observation with mixed-type variables. To calculate distance \textcolor{black}{between observations using a categorical-only metric}, we first homogenize the data using discretization\textcolor{black}{.} The $k^\textup{th}$ continuous variable \textcolor{black}{of} $\mathbf{x}_i$ \textcolor{black}{is divided} into $C=\{1,2,\ldots,c_k\}$ ordered categories of \textcolor{black}{disjoint intervals} $\mathcal{Z}_1, \mathcal{Z}_2, \ldots, \mathcal{Z}_{c_k}$.\textcolor{black}{Then, we define new order categorical variable such that $\mathbf{z}_k=\{z_{i,k}=c|x_{i,k}\in \mathcal{Z}_c\}$} and replace each value $x_{i,k}$ with $z_{i,k}$. Discretization is often useful in cases where the data is highly skewed. However, it leads to a loss of information, and choosing an optimal interval width can be challenging and affect the analysis results.\\

\textcolor{black}{Another method for coercing continuous data to categorical is d}ummy-coding \textcolor{black}{that involves representing binned categorical variables as binary (0 or 1) variables. For each category in $C$, a corresponding binary variable is created $x_{i,h,m}$, where the binary variable assumes value 1 if $\mathbf{x}_i$ and $\mathbf{x}_j$ are both in category $c_{h,m}$, and 0 otherwise.} The distance \textcolor{black}{between any two dummy coded observations} $d(\mathbf{x}_i, \mathbf{x}_j)$ \textcolor{black}{can be} calculated using any binary distance metric (for more information, see Choi et al., 2010). Dummy-coding has the advantage of preserving all the information in the categorical variable and ease of interpretation. However, such an approach increase\textcolor{black}{s} the dimensions of the feature space \textcolor{black}{and ignores the ordered values of continuous scales}. \textcolor{black}{Foss et al. (2019) illustrated the inadequacy of dummy coding, noting that the expectation of the interval scale variable is always greater than 1, while the expectation from the categorical is always less than 1. This means that the choice of coding can lead to different interpretations of the data and may affect the analysis results.}\\

\textcolor{black}{The coercion of categorical data to continuous is typically conducted through an ordered numerical assignment such as converting Likert scale data to numerical. Then, the differences are calculated using any continuous metric, such as Euclidean distance. Additionally, continuous data may} need to be scaled \textcolor{black}{as not to over or under weigh the contribution of individual continuous variables}, and the choice of scaling also affects \textcolor{black}{distance calculations and thus} clustering performance. Hennig et al. (2015) noted that distance-based clustering methods are not invariant to affine transformations, and Foss et al. (2016) showed that the choice of scaling can affect clustering performance. \\

Various mixed\textcolor{black}{-type} distance metrics do not require the homogenization or scaling of the data. The quadratic distance proposed by Lindsay et al. (2008) extends the chi-squared measures of distance between two distributions and requires the choice of a nonnegative definite kernel. De Leon \& Carriere (2005) use a general mixed-data model to define the distance between two populations of mixed unordered categorical, ordered categorical, and interval scale data. Krzanowski (1983) proposes a distance based on Matusita's distance, as mixtures of these location models and generalizations are not identifiable without further conditions on some of the parameters. Recently, van de Welden et al., (2023) introduced a framework that allows for an implementation of distances between observations that can be extended to many existing distances for categorical variables. \textcolor{black}{Modha and Spangler (2003) propose a method similar to $k-$prototypes that includes estimating a suitable weight that scales the relative contribution of the interval and categorical variables. However, the brute-force search to cluster repeatedly for a range of values for the weight that minimizes its objective function is computationally exhaustive.} \\

The metrics \textcolor{black}{shown} in Table \ref{tab:distances} will be used as benchmarks for the analysis of metrics herein. \textcolor{black}{For any arbitrary variable $l$, denote $w_{i,j,l} = 0$ if variable $l$ has missing data, otherwise $w_{i,j,l} = 1$. Denote $\textcolor{black}{\mathds{1}}_{i,j,l} \equiv \mathds{1}_c(x_{i,l} = x_{j,l})$ for categorical variables, where $\mathds{1}_{i,j,l} = 1$ if the two observations for the $l$th variables are the same, and $0$ otherwise.} Gower's distance (Gower, 1971) is a common hybrid distance function that calculates the distance between two vectors of the same length. It uses a weighted combination of interval and categorical distances, where the categorical distance is based on whether the categories match or not, and the interval distance is scaled based on the range of the variable. The user-specified weights for each variable may lead to intractable solutions and varying results based on the data. \textcolor{black}{The} $k-$prototypes \textcolor{black}{algorithm} (Huang, 1998) \textcolor{black}{utilizes} another hybrid distance technique that uses a similar approach to Gower's distance, except the squared Euclidean distance is used for the interval scale variables. Unlike Gower's distance, it does not require variable-specific weights, but rather a single weight used for the entire categorical contribution of the distance function. The Podani distance metric (1999) extends Gower's general coefficient of similarity to ordinal variables, while the Wishart (2003) metric is similar to the Podani metric, except it makes use of the sample standard deviation for continuous variables, rather than the range of the continuous variables. \\

\begin{table}[H]
    \centering
    \begin{tabular}{c c}
    Metric & Definition \\ 
    \hline \hline
    Gower (1971):           & $d(\mathbf{x}_i,\mathbf{x}_j) = 1 - s_{i,j};$  \\
       & $s_{i,j} = \frac{\sum_{l=1}^pw_{i,j,l}\textcolor{black}{s}_{i,j,l}}{\sum_{l=1}^pw_{i,j,l}}$ \\
       & $\textcolor{black}{s_{i,j,l} = 1 - \frac{|x_{i,l} - x_{j,l}|}{\max_k(x_{k,l}) - \min_k(x_{k,l})}}$ \\
    \hline
    Huang (\textcolor{black}{1998}):           & $d(\mathbf{x}_i,\mathbf{x}_j) = \sum_{l=1}^{p_c}(x_{i,l} - x_{j,l})^2 +$ \\ & $+ \gamma\sum_{l=p_c + 1}^{p_u + p_o}\textcolor{black}{\mathds{1}}_c(x_{i,l}=x_{j,l});$ \\
                    & $\gamma = \frac{\sum_{r=1}^{p_c}s_r^2}{p_c}$ \\
    \hline
    Podani (1999):          & $d(\mathbf{x}_i,\mathbf{x}_j) = \sqrt{\sum_{l=1}^p w_{i,j,l}\left(\frac{x_{i,l}-x_{j,l}}{\textcolor{black}{s}_{i,j,l}}\right)^2};$ \\
     &  $\textcolor{black}{s}_{i,j,l} = \max_k(x_{k,l}) - \min_k(x_{k,l})$ (if $l$ is continuous)  \\
    \hline 
    Wishart (2003):         & $d(\mathbf{x}_i,\mathbf{x}_j) = \sqrt{\sum_{\textcolor{black}{r}=1}^p w_{i,j,\textcolor{black}{r}}\left(\frac{x_{i,\textcolor{black}{r}}-x_{j,\textcolor{black}{r}}}{\textcolor{black}{s_r}}\right)^2};$ \\
    & $\textcolor{black}{s_r} :=$ sample standard deviation of $l$th variable (if $l$ is continuous)\\
    \end{tabular}
    \caption{\textcolor{black}{Common} mixed\textcolor{black}{-type} distance metrics.}
    \label{tab:distances}
\end{table}

\section{Kernel Density Estimation and Bandwidth Selection Procedures} \label{sec:kernel}

Kernel functions are weighting functions that can map data points from a high-dimensional sample space to a low-dimensional space. \textcolor{black}{Kernel functions are non-negative real-valued functions, integrate or sum to 1, and often assumed to be symmetric.} We denote the kernel functions specific to datatypes as $K$, $L$, and $l$ for continuous, unordered and ordered categorical variables, respectively. For each kernel, we denote bandwidths associated with the kernel functions as $\boldsymbol{\lambda} \equiv \{\textcolor{black}{\boldsymbol{\lambda}}^c, \textcolor{black}{\boldsymbol{\lambda}}^u, \textcolor{black}{\boldsymbol{\lambda}}^o\}$ where $\{\textcolor{black}{\boldsymbol{\lambda}}^c\} \equiv \{\lambda_{i}\}_{i=1}^{p_c}, \{\textcolor{black}{\boldsymbol{\lambda}}^u\} \equiv \{\lambda_{i}\}_{i=p_c+1}^{p_c+p_u},$ and $\{\textcolor{black}{\boldsymbol{\lambda}}^o\} \equiv \{\lambda_{i}\}_{i=p_c+p_u+1}^{p}$.\\

\textcolor{black}{There exist many c}ommon kernel functions used in the smoothing literature\textcolor{black}{, such as the} Gaussian kernel for continuous variables \textcolor{black}{given by}
\begin{eqnarray}\label{eq:gauss}
 \textcolor{black}{k(x_i,x,\lambda^c) = \frac{1}{\sqrt{2\pi}}e^{\frac{-(x_i-x)^2}{2(\lambda^c)^2}},}
\end{eqnarray}
\textcolor{black}{where $\lambda^c >0$. The Epanechnikov kernel (Epanechnikov, 1969) given by}
\begin{eqnarray}\label{eq:epanech}
\textcolor{black}{k(z) = k(x_i,x,\lambda^c) = \frac{3}{4}\left(1-\frac{(x_i-x)^2}{(\lambda^c)^2}\right)\textcolor{black}{\mathds{1}}\left(\left|\frac{x_i-x}{\lambda^c}\right| \leq 1\right),} %
\end{eqnarray}
\textcolor{black}{where $\lambda^c >0$. F}or unordered categorical variables\textcolor{black}{, we use an Aitken kernel \textcolor{black}{(Li and Racine, 2023)} given by}
\begin{eqnarray} \label{eq:unordered}
\textcolor{black}{L(x_i,x,\lambda^u) = \begin{cases} 1, & x_i = x, \\ \lambda^u, & x_i \neq x, \end{cases}}
\end{eqnarray}
\textcolor{black}{where $\lambda^u\in [0,1]$. F}or ordered categorical \textcolor{black}{variables, a Wang \& van Ryzin kernel (Wang \& van Ryzin, 2008) given by}
\begin{eqnarray}\label{eq:ordered}
\textcolor{black}{l(x_i,x,\lambda^o) = \begin{cases} 1 - \lambda^o, & x_i = x, \\ \frac{1}{2}\left(1-\lambda^o\right)\left(\lambda^o\right)^{|X_i-x|}, & x_i \ne x ,\end{cases}}
\end{eqnarray}
\textcolor{black}{where $\lambda^o\in [0,1]$. }A mixed-type joint kernel function between a random vector $\mathbf{x}_j \equiv \{\mathbf{x}_j^c,\mathbf{x}_j^u,\mathbf{x}_j^o\}$ and an arbitrary point $\mathbf{x}$ is written as
\begin{equation}\label{eq:mlcv1}
\textcolor{black}{K}(\mathbf{x}_i,\mathbf{x})= \prod_{k=1}^{p_c}\frac{1}{\lambda_k}k\left(\frac{\textcolor{black}{x_{i,k}}^c-x_{k}^c}{\lambda_{k}}\right)\prod_{k=\textcolor{black}{p_c+}1}^{\textcolor{black}{p_c+p_u}}L\left(x_{i,k}^u,\textcolor{black}{x}_{k}^u,\lambda_k\right)\times \cdots\times \prod_{k=\textcolor{black}{p_c+p_u+1}}^{\textcolor{black}{p}}l\left(x_{i,k}^o,\textcolor{black}{x}_{k}^o,\lambda_k\right).
\end{equation}

\textcolor{black}{These kernel functions can be used for probability density estimation, such as the Rozenblatt-Parzen density estimator (\cite{rosenblatt1956remarks,parzen1962estimation}) $\widehat{p}(\mathbf{x})=\frac{1}{n\lambda_1\lambda_2\cdots\lambda_{p_c}}\sum_{i=1}^nK(\mathbf{x}_i,\mathbf{x})$ converges in probability to the underlying density function $p(x)$ under the assumption that as $n\to\infty$, then $\boldsymbol{\lambda}\to \mathbf{0}$ and $n\lambda_1\lambda_2\cdots\lambda_{p_c}\to\infty$.} Optimal bandwidth selection methods are designed to preserve estimator convergence while having several other desirable properties, including smoothing out irrelevant variables (Loader, 1999). There is a wide range of methods for optimal bandwidth selection, including Akaike Information Criterion (Hurvich et al., 1998), Least Squares Cross-Validation (Sain et al., 1994), Rule of Thumb (Silverman, 1986), and Maximum-Likelihood Cross-Validation (MLCV) (Hall 1981). \textcolor{black}{T}he MLCV objective function to be minimized is 
\begin{eqnarray}\label{eq:cvml}
CV(\boldsymbol{\lambda}) = \sum_{i=1}^n\ln\left(\frac{1}{(n-1)}\sum_{j=1, j \ne i}^n\mathcal{L}_{\boldsymbol{\lambda}}(x_{i,k},x_{j,k})\right) = \sum_{i=1}^n\ln\left(\hat{\mathcal{L}}_{-i}(x_i)\right),\end{eqnarray} where $\hat{\mathcal{L}}_{-i}(x_i)$ is the leave-one-out estimator of $\mathcal{L}_{\boldsymbol{\lambda}}(\cdot)$ in Equation (\ref{eq:mlcv1}). \textcolor{black}{These kernels and cross-validation approaches can be adapted to similarity functions, which are used for the kernel distance metric in this paper.}

\section{\textcolor{black}{Similarity Functions and} Mixed\textcolor{black}{-type} Kernel Distances} \label{sec:methods}
\textcolor{black}{Consider a} real-valued function $\mathcal{L}(\mathbf{x}_{\textcolor{black}{i}}, \mathbf{x}_{\textcolor{black}{j}})$ on the Cartesian product $X \times X$\textcolor{black}{. $\mathcal{L}$} is a similarity function if, for any points $\mathbf{x}_{\textcolor{black}{i}}, \mathbf{x}_{\textcolor{black}{j}}, \mathbf{x}_{\textcolor{black}{k}} \in X$, it satisfies four conditions (Chen et al., 2009): 
\begin{itemize}
\item[(S1)] Symmetry: $\mathcal{L}(\mathbf{x}_{\textcolor{black}{i}},\mathbf{x}_{\textcolor{black}{j}}) = \mathcal{L}(\mathbf{x}_{\textcolor{black}{j}},\mathbf{x}_{\textcolor{black}{i}})$, 
\item[(S2)] Indiscernible: $\mathcal{L}(\mathbf{x}_{\textcolor{black}{i}},\mathbf{x}_{\textcolor{black}{j}}) = \mathcal{L}(\mathbf{x}_{\textcolor{black}{i}},\mathbf{x}_{\textcolor{black}{i}}) = \mathcal{L}(\mathbf{x}_{\textcolor{black}{j}},\mathbf{x}_{\textcolor{black}{j}}) \iff \mathbf{x}_{\textcolor{black}{i}} = \mathbf{x}_{\textcolor{black}{j}}$, 
\item[(S3)] Nonnegative self-similarity: $\mathcal{L}(\mathbf{x}_{\textcolor{black}{i}},\mathbf{x}_{\textcolor{black}{i}}) \geq \mathcal{L}(\mathbf{x}_{\textcolor{black}{i}},\mathbf{x}_{\textcolor{black}{j}}) \geq 0$,
\item[(S4)] Similarity triangle inequality: $\mathcal{L}(\mathbf{x}_{\textcolor{black}{i}},\mathbf{x}_{\textcolor{black}{j}})+\mathcal{L}(\mathbf{x}_{\textcolor{black}{j}},\mathbf{x}_{\textcolor{black}{k}}) \leq \mathcal{L}(\mathbf{x}_{\textcolor{black}{i}},\mathbf{x}_{\textcolor{black}{k}}) + \mathcal{L}(\mathbf{x}_{\textcolor{black}{j}},\mathbf{x}_{\textcolor{black}{j}})$.
\end{itemize}
\textcolor{black}{Let $\mathcal{L}(\cdot)$ be a similarity function that maps two $p$-dimensional data vectors to the real numbers with the additional property that as the difference between two observations $\mathbf{x}_i$ and $\mathbf{x}_j$ increases, $\mathcal{L}(\mathbf{x}_i,\mathbf{x}_j)$ decreases.\\} 

\textcolor{black}{The definition of norm and distance between two vectors is defined by the choice of kernel similarity function. For example, a Euclidean norm is positive, definite, and symmetric, and yields the similarity function $\mathcal{L}_{L_2}(\mathbf{x}_{\textcolor{black}{i}}, \mathbf{x}_{\textcolor{black}{j}})=1-\sum_{k=1}^p(x_{i,k}-x_{j,k})^2$. The Gaussian kernel function defined in Equation (\ref{eq:gauss}) can be written in terms of a non-linear transformation of  $\mathcal{L}_{L_2}$ as $\mathcal{L}_{gauss}(x_i,x_j):=k\left(\frac{x_i-x_j}{\lambda_{k}}\right)=\frac{1}{\sqrt{2\pi}}\mbox{exp}\left(-\frac{1}{2}\left(\frac{x_i-x_j}{\lambda_{k}}\right)^2\right)$. Each of these kernels satisfy the necessary conditions of positive, definite and symmetric kernel functions (\cite{bekka2008kazhdan}) and satisfy the properties (S1)-(S4) to be similarity functions (Joshi et al., 2011; Phillips \& Venkatasubramanian, 2011; J{\"a}kel et al., 2008). The Epanechnikov kernel function in Equation (\ref{eq:epanech}) can be viewed as a scaled and truncated version of $\mathcal{L}_{L_2}$ by using $z=\frac{x_{i,k}-x_{j,k}}{h}$. As an example for continuous kernels, we show that the Gaussian is a similarity functions.}\\

\textcolor{black}{\noindent\textit{Lemma 1:} The Gaussian kernel is a similarity function.}\\
\textcolor{black}{\textit{Proof}:}
\textcolor{black}{We demonstrate that the Gaussian kernel $k(x_i,x_j,\lambda^c) = \frac{1}{\sqrt{2\pi}}e^{\frac{-(x_i-x_j)^2}{2(\lambda^c)^2}}$ satisfies properties (S1)-(S4) of a similarity function $\forall \: \lambda^c >0$.\\
(S1): Since $(x_i-x_j)^2=(x_j-x_i)^2$, then $k(x_i,x_j,\lambda^c)=k(x_j,x_i,\lambda^c)$. \\
(S2): ($\Rightarrow$) Let $k(x_i,x_j,\lambda^c) = k(x_i,x_i,\lambda^c) = k(x_j,x_j,\lambda^c)=\frac{1}{\sqrt{2\pi}}$ but assume that $x_i\neq x_j$. Considering that $k(x_i,x_i,\lambda^c)$, then $k(x_j,x_j,\lambda^c)=\frac{1}{\sqrt{2\pi}}$ which implies $(x_i-x_j)^2=0$. However, $(x_i-x_j)^2>0$ by the assumption $x_i\neq x_j$ which is a contradiction. Thus, $k(x_i,x_j,\lambda^c) = k(x_i,x_i,\lambda^c) = k(x_j,x_j,\lambda^c)=\frac{1}{\sqrt{2\pi}} \Rightarrow x_i = x_j$. ($\Leftarrow$) Let $x_i = x_j$, then by (S1) $k(x_i,x_j,\lambda^c) = k(x_i,x_i,\lambda^c) = k(x_j,x_j,\lambda^c) = \frac{1}{\sqrt{2\pi}}$. \\
(S3): First, $k(x_i,x_i,\lambda^c) = \frac{1}{\sqrt{2\pi}} > 0$. Assuming $x_i\neq x_j$, then $\frac{-(x_i-x_j)^2}{2(\lambda^c)^2}<0$ and $e^\frac{-(x_i-x_j)^2}{2(\lambda^c)^2}< 1$, and $k(x_i,x_i,\lambda^c) > k(x_i,x_j,\lambda^c)$. If $x_i=x_j$, then by (S2) $k(x_i,x_i,\lambda^c) = k(x_i,x_j,\lambda^c)$. Thus, $k(x_i,x_i,\lambda^c) \geq k(x_i,x_j,\lambda^c) \geq 0$.\\
(S4): If either $x_i = x_j$ or $x_j = x_k$, then the inequality holds by (S1)-(S3). Consider the case when $x_i \ne x_j$ and $x_j \ne x_k$. The Euclidean norm triangle inequality gives
\begin{eqnarray}
& (x_i - x_k)^2  \leq (x_i - x_j)^2 + (x_j - x_k)^2, \nonumber \\
\implies & 2 - \left(\frac{1}{2(\lambda^c)^2}(x_i - x_j)^2 + \frac{1}{2(\lambda^c)^2}(x_j - x_k)^2\right) \leq 2 - \frac{1}{2(\lambda^c)^2}(x_i - x_k)^2 \label{eq:gaussianSim1}
\end{eqnarray}
Consider $e^{-x} \geq 1 - x$, thus $e^{\frac{-(x_i-x_j)^2}{2(\lambda^c)^2}} \geq 1 - \frac{1}{2(\lambda^c)^2}(x_i - x_j)^2$. Using this inequality gives 
\begin{eqnarray} \label{eq:gaussianSim2}
    e^{\frac{-(x_i-x_j)^2}{2(\lambda^c)^2}} + e^{\frac{-(x_j-x_k)^2}{2(\lambda^c)^2}} \geq 2 - \left(\frac{1}{2(\lambda^c)^2}(x_i - x_j)^2 + \frac{1}{2(\lambda^c)^2}(x_j - x_k)^2\right) 
\end{eqnarray}
and 
\begin{eqnarray} \label{eq:gaussianSim3}
   1+ e^{\frac{-(x_i-x_k)^2}{2(\lambda^c)^2}} \geq 2 - \frac{1}{2(\lambda^c)^2}(x_i - x_k)^2
\end{eqnarray}
Inserting Equations (\ref{eq:gaussianSim2}) and (\ref{eq:gaussianSim3}) into (\ref{eq:gaussianSim1}) yields
\begin{eqnarray*}
 & e^{\frac{-(x_i-x_j)^2}{2(\lambda^c)^2}} + e^{\frac{-(x_j-x_k)^2}{2(\lambda^c)^2}} \leq e^{\frac{-(x_i-x_k)^2}{2(\lambda^c)^2}} + 1 \\
 \implies & \frac{1}{\sqrt{2\pi}}e^{\frac{-(x_i-x_j)^2}{2(\lambda^c)^2}} + \frac{1}{\sqrt{2\pi}}e^{\frac{-(x_j-x_k)^2}{2(\lambda^c)^2}}  \leq \frac{1}{\sqrt{2\pi}}e^{\frac{-(x_i-x_k)^2}{2(\lambda^c)^2}} + \frac{1}{\sqrt{2\pi}} \\
 \implies & k(x_i, x_j, \lambda^c) + k(x_j,x_k,\lambda^c) \leq k(x_i,x_k,\lambda^c) + k(x_j,x_j,\lambda^c)   
\end{eqnarray*}
which is (S4) and thus the Gaussian kernel is a similarity function.}\; \; \; \; \; \; \; \; \; \; \; \; \; \; \; \; \; \; \; \; \; \; \; \; \; \; \; \; \; \; \; \; \; \; \; \; \; \; \; \; \; \; \; \; \; \textcolor{black}{$\square$}
\textcolor{black}{Similar arguments can be made to show that an Epanechnikov kernel function is a similarity function, which we omit for brevity. For categorical kernels, there is far less literature support for their usage as similarity functions. We show that Equations (\ref{eq:unordered}) and (\ref{eq:ordered}) are similarity functions.}
\textcolor{black}{\noindent\textit{Lemma 2:} The Aitken unordered kernel is a similarity function.}\\
\textcolor{black}{\textit{Proof}:}
\textcolor{black}{We demonstrate that the Aitken Kernel satisfies all properties of a similarity function (S1)-(S4).}
\textcolor{black}{(S1): By the definition of symmetry for this kernel function, (S1) is satisfied. \\
(S2): Suppose $L(x_i,x_j,\lambda^u) = L(x_i,x_i,\lambda^u) = L(x_j,x_j,\lambda^u)$. Then $k(\cdot) = 1$, by definition. Conversely, if $x_i = x_j$, then $L(x_i,x_j,\lambda^u) = L(x_i,x_i,\lambda^u) = L(x_j,x_j,\lambda^u) = 1$. \\
(S3): First, $L(x_i,x_i,\lambda^u) = 1 \geq 0$. Then, $L(x_i,x_j,\lambda^u) = \lambda^u \leq 1$, based on the bounds of $\lambda^u$. Thus, $L(x_i,x_i,\lambda^u) \geq L(x_i,x_j,\lambda^u) \geq 0$.\\
(S4): If $x_i = x_j$ or $x_j = x_k$, the result is trivial. Assume $x_i \ne x_j$ and $x_j \ne x_k$. Then,
\begin{align*}
    L(x_i, x_j, \lambda^u) + L(x_j,x_k,\lambda^u) & \leq L(x_i,x_k,\lambda^u) + L(x_j,x_j,\lambda^u) \\
    \implies \lambda^u + \lambda^u & \leq \lambda^u + 1 \\
    \implies \lambda^u & \leq 1, \\
\end{align*}
    which is true based on the bounds of $\lambda^u$.}\; \; \; \; \; \; \; \; \; \; \; \; \; \; \; \; \; \; \; \; \; \; \; \; \; \; \; \; \; \; \; \; \; \; \; \; \; \; \; \; \; \; \; \; \; \textcolor{black}{$\square$}

\textcolor{black}{\noindent\textit{Lemma 3:} The  Wang \& van Ryzin ordered kernel is a similarity function.}\\
\textcolor{black}{\textit{Proof}:}
\textcolor{black}{We demonstrate that the  Wang \& van Ryzin Kernel satisfies all properties of a similarity function (S1)-(S4).}
\textcolor{black}{By definition, (S1) and (S2) are satisfied. \\
(S3): First, $l(x_i,x_i,\lambda^o) = 1 - \lambda^o \in [0,1]$. Then, $l(x_i,x_j,\lambda^o) = \frac{1}{2}(1-\lambda^o)(\lambda^o)^{|x_i - x_j|}$. To show $1 - \lambda^o \geq \frac{1}{2}(1-\lambda^o)(\lambda^o)^{|x_i - x_j|}$, divide both sides by $1-\lambda^o$ to see $1 \geq \frac{1}{2}(\lambda^o)^{|x_i - x_j|}$ (note that we can divide by $1-\lambda^o$, since if $\lambda^o = 1$, we have $0 \geq 0$ which holds). Since $|x_i - x_j|$ is a positive integer, and since $\lambda^o \in [0,1]$, $(\lambda^o)^{|x_i - x_j|} \in [0,1]$, and the inequality follows.\\
(S4): If $x_i = x_j$ or $x_j = x_k$, the result is trivial. Assume $x_i \ne x_j$ and $x_j \ne x_k$. Then,
\begin{align*}
    l(x_i, x_j, \lambda^o) + l(x_j,x_k,\lambda^o) & \leq l(x_i,x_k,\lambda^o) + l(x_j,x_j,\lambda^o) \\
    \implies \frac{1}{2}(1-\lambda^o)(\lambda^o)^{|x_i - x_j|} + \frac{1}{2}(1-\lambda^o)(\lambda^o)^{|x_j - x_k|}  & \leq \frac{1}{2}(1-\lambda^o)(\lambda^o)^{|x_i - x_k|} + (1 - \lambda^o)  \\
    \implies \frac{1}{2}\left((\lambda^o)^{|x_i-x_j|} + (\lambda^o)^{|x_j - x_k|}\right) & \leq \frac{1}{2}(\lambda^o)^{|x_i-x_k|} + 1. \\
\end{align*}
    Now, $\frac{1}{2}\left((\lambda^o)^{|x_i-x_j|} + (\lambda^o)^{|x_j - x_k|}\right) \leq \max\left\{(\lambda^o)^{|x_i-x_j|},(\lambda^o)^{|x_j - x_k|}\right\}$, so it will be sufficient to show $$\max\left\{(\lambda^o)^{|x_i-x_j|},(\lambda^o)^{|x_j - x_k|}\right\} \leq \frac{1}{2}(\lambda^o)^{|x_i-x_k|}+ 1.$$ If $|x_i - x_j| \leq |x_j - x_k|$, then $\max\left\{(\lambda^o)^{|x_i-x_j|},(\lambda^o)^{|x_j - x_k|}\right\} = (\lambda^o)^{|x_j - x_k|}$, so we show $(\lambda^o)^{|x_j - x_k|} \leq \frac{1}{2}(\lambda^o)^{|x_i - x_k|} + 1$, which is clear since $(\lambda^o)^{|x_i - x_k|} \in [0,1]$. A similar argument holds if $|x_i - x_j| > |x_j - x_k|$, and the result follows immediately.}\; \; \; \; \; \; \; \; \; \; \; \; \; \; \; \; \; \; \; \; \; \; \; \; \; \; \; \; \; \; \; \; \; \; \; \; \; \; \; \; \; \; \; \; \; \; \; \; \; \; \; \; \; \; \; \; \; \; \; \textcolor{black}{$\square$}

To transform kernel similarities into distances, we extend the metric described in Phillips and Venkatasubramanian (2011) to the multivariate setting, which uses a well-defined kernel function to measure similarity between points $\mathbf{x}_{\textcolor{black}{i}}$ and $\mathbf{x}_{\textcolor{black}{j}}$. The distance between $\mathbf{x}_{\textcolor{black}{i}}$ and $\mathbf{x}_{\textcolor{black}{j}}$ is then defined as \begin{equation}\label{eq:dist} d(\mathbf{x}_{\textcolor{black}{i}}, \mathbf{x}_{\textcolor{black}{j}}) = \mathcal{L}(\mathbf{x}_{\textcolor{black}{i}}, \mathbf{x}_{\textcolor{black}{i}}) + \mathcal{L}(\mathbf{x}_{\textcolor{black}{j}}, \mathbf{x}_{\textcolor{black}{j}}) - \mathcal{L}(\mathbf{x}_{\textcolor{black}{i}}, \mathbf{x}_{\textcolor{black}{j}}) - \mathcal{L}(\mathbf{x}_{\textcolor{black}{j}}, \mathbf{x}_{\textcolor{black}{i}}). \end{equation} If a symmetric kernel function is elected, the formula reduces to $d(\mathbf{x}_{\textcolor{black}{i}}, \mathbf{x}_{\textcolor{black}{j}}) = \mathcal{L}(\mathbf{x}_{\textcolor{black}{i}}, \mathbf{x}_{\textcolor{black}{i}}) + \mathcal{L}(\mathbf{x}_{\textcolor{black}{j}}, \mathbf{x}_{\textcolor{black}{j}}) - 2\mathcal{L}(\mathbf{x}_{\textcolor{black}{i}}, \mathbf{x}_{\textcolor{black}{j}})$. Equation (\ref{eq:dist}) represents the difference between the self-similarities of the two points and their cross-similarity. The multiplicative factor of two ensures that the distance between an object and itself equals zero and satisfies the identity of indiscernibles. \\

\noindent\textit{Theorem 1:} \textcolor{black}{Equation (\ref{eq:dist}) that satisfies (S1)-(S4)} is a well-defined distance metric \textcolor{black}{where it} satisfies the following distant metric properties (Chen et al., 2009): \\
\textit{Proof}: \textcolor{black}{(D1)} Nonnegativity ($d(\mathbf{x}_{\textcolor{black}{i}},\mathbf{x}_{\textcolor{black}{j}}) \geq 0)$: note by (S3) that
$\mathcal{L}(\mathbf{x}_{\textcolor{black}{i}},\mathbf{x}_{\textcolor{black}{i}}) - \mathcal{L}(\mathbf{x}_{\textcolor{black}{i}},\mathbf{x}_{\textcolor{black}{j}}) \geq 0$  and $\mathcal{L}(\mathbf{x}_{\textcolor{black}{j}},\mathbf{x}_{\textcolor{black}{j}}) - \mathcal{L}(\mathbf{x}_{\textcolor{black}{j}},\mathbf{x}_{\textcolor{black}{i}}) \geq 0$. Adding yields
$\mathcal{L}(\mathbf{x}_{\textcolor{black}{i}},\mathbf{x}_{\textcolor{black}{i}}) + \mathcal{L}(\mathbf{x}_{\textcolor{black}{j}},\mathbf{x}_{\textcolor{black}{j}}) - \mathcal{L}(\mathbf{x}_{\textcolor{black}{i}},\mathbf{x}_{\textcolor{black}{j}}) - \mathcal{L}(\mathbf{x}_{\textcolor{black}{j}},\mathbf{x}_{\textcolor{black}{i}}) \geq 0$ and thus, $d(\mathbf{x}_{\textcolor{black}{i}},\mathbf{x}_{\textcolor{black}{j}}) \geq 0$. 

\noindent \textcolor{black}{(D2)} Symmetry ($d(\mathbf{x}_{\textcolor{black}{i}},\mathbf{x}_{\textcolor{black}{j}}) = d(\mathbf{x}_{\textcolor{black}{j}},\mathbf{x}_{\textcolor{black}{i}})$): note $d(\mathbf{x}_{\textcolor{black}{i}},\mathbf{x}_{\textcolor{black}{j}}) = \mathcal{L}(\mathbf{x}_{\textcolor{black}{i}},\mathbf{x}_{\textcolor{black}{i}}) + \mathcal{L}(\mathbf{x}_{\textcolor{black}{j}},\mathbf{x}_{\textcolor{black}{j}}) - \mathcal{L}(\mathbf{x}_{\textcolor{black}{i}},\mathbf{x}_{\textcolor{black}{j}}) - \mathcal{L}(\mathbf{x}_{\textcolor{black}{j}},\mathbf{x}_{\textcolor{black}{i}})  
= \mathcal{L}(\mathbf{x}_{\textcolor{black}{j}},\mathbf{x}_{\textcolor{black}{j}}) + \mathcal{L}(\mathbf{x}_{\textcolor{black}{i}},\mathbf{x}_{\textcolor{black}{i}}) - \mathcal{L}(\mathbf{x}_{\textcolor{black}{j}},\mathbf{x}_{\textcolor{black}{i}}) - \mathcal{L}(\mathbf{x}_{\textcolor{black}{i}},\mathbf{x}_{\textcolor{black}{j}}) = d(\mathbf{x}_{\textcolor{black}{j}},\mathbf{x}_{\textcolor{black}{i}})$ 

\noindent \textcolor{black}{(D3)}  Identity of indescernibles ($d(\mathbf{x}_{\textcolor{black}{i}}, \mathbf{x}_{\textcolor{black}{j}}) = 0 \iff \mathbf{x}_{\textcolor{black}{i}}=\mathbf{x}_{\textcolor{black}{j}}$): suppose that $d(\mathbf{x}_{\textcolor{black}{i}},\mathbf{x}_{\textcolor{black}{j}}) = 0$, thus
$\mathcal{L}(\mathbf{x}_{\textcolor{black}{i}},\mathbf{x}_{\textcolor{black}{i}}) + \mathcal{L}(\mathbf{x}_{\textcolor{black}{j}},\mathbf{x}_{\textcolor{black}{j}}) - \mathcal{L}(\mathbf{x}_{\textcolor{black}{i}},\mathbf{x}_{\textcolor{black}{j}}) - \mathcal{L}(\mathbf{x}_{\textcolor{black}{j}},\mathbf{x}_{\textcolor{black}{i}}) = 0$, implying $\mathcal{L}(\mathbf{x}_{\textcolor{black}{i}},\mathbf{x}_{\textcolor{black}{i}}) + \mathcal{L}(\mathbf{x}_{\textcolor{black}{j}},\mathbf{x}_{\textcolor{black}{j}}) = \mathcal{L}(\mathbf{x}_{\textcolor{black}{i}},\mathbf{x}_{\textcolor{black}{j}}) + \mathcal{L}(\mathbf{x}_{\textcolor{black}{j}},\mathbf{x}_{\textcolor{black}{i}}) $ which is true if and only if $\mathbf{x}_{\textcolor{black}{i}} = \mathbf{x}_{\textcolor{black}{j}}$ or $\mathbf{x}_{\textcolor{black}{j}}=\mathbf{x}_{\textcolor{black}{i}}$ by (S2). Conversely, suppose $\mathbf{x}_{\textcolor{black}{j}}=\mathbf{x}_{\textcolor{black}{i}}$, then
$d(\mathbf{x}_{\textcolor{black}{i}},\mathbf{x}_{\textcolor{black}{i}}) = \mathcal{L}(\mathbf{x}_{\textcolor{black}{i}},\mathbf{x}_{\textcolor{black}{i}}) + \mathcal{L}(\mathbf{x}_{\textcolor{black}{i}},\mathbf{x}_{\textcolor{black}{i}}) - \mathcal{L}(\mathbf{x}_{\textcolor{black}{i}},\mathbf{x}_{\textcolor{black}{i}}) - \mathcal{L}(\mathbf{x}_{\textcolor{black}{i}},\mathbf{x}_{\textcolor{black}{i}}) = 2\mathcal{L}(\mathbf{x}_{\textcolor{black}{i}},\mathbf{x}_{\textcolor{black}{i}}) - 2\mathcal{L}(\mathbf{x}_{\textcolor{black}{i}},\mathbf{x}_{\textcolor{black}{i}}) = 0$ 

\noindent \textcolor{black}{(D4)} Triangle inequality ($d(\mathbf{x}_{\textcolor{black}{i}}, \mathbf{x}_{\textcolor{black}{k}}) \leq d(\mathbf{x}_{\textcolor{black}{i}}, \mathbf{x}_{\textcolor{black}{j}}) + d(\mathbf{x}_{\textcolor{black}{j}}, \mathbf{x}_{\textcolor{black}{k}})$): note by (S4) that $\mathcal{L}(\mathbf{x}_{\textcolor{black}{i}},\mathbf{x}_{\textcolor{black}{j}}) + \mathcal{L}(\mathbf{x}_{\textcolor{black}{j}},\mathbf{x}_{\textcolor{black}{k}}) \leq \mathcal{L}(\mathbf{x}_{\textcolor{black}{i}},\mathbf{x}_{\textcolor{black}{k}}) + \mathcal{L}(\mathbf{x}_{\textcolor{black}{j}},\mathbf{x}_{\textcolor{black}{j}})$, and $\mathcal{L}(\mathbf{x}_{\textcolor{black}{k}},\mathbf{x}_{\textcolor{black}{j}}) + \mathcal{L}(\mathbf{x}_{\textcolor{black}{j}},\mathbf{x}_{\textcolor{black}{i}}) \leq \mathcal{L}(\mathbf{x}_{\textcolor{black}{k}},\mathbf{x}_{\textcolor{black}{i}}) + \mathcal{L}(\mathbf{x}_{\textcolor{black}{j}},\mathbf{x}_{\textcolor{black}{j}})$. Then, \\$d(\mathbf{x}_{\textcolor{black}{i}},\mathbf{x}_{\textcolor{black}{k}}) = \mathcal{L}(\mathbf{x}_{\textcolor{black}{i}},\mathbf{x}_{\textcolor{black}{i}}) + \mathcal{L}(\mathbf{x}_{\textcolor{black}{k}},\mathbf{x}_{\textcolor{black}{k}}) - \mathcal{L}(\mathbf{x}_{\textcolor{black}{i}},\mathbf{x}_{\textcolor{black}{k}}) - \mathcal{L}(\mathbf{x}_{\textcolor{black}{k}},\mathbf{x}_{\textcolor{black}{i}})
\leq \mathcal{L}(\mathbf{x}_{\textcolor{black}{i}},\mathbf{x}_{\textcolor{black}{i}}) + \mathcal{L}(\mathbf{x}_{\textcolor{black}{k}},\mathbf{x}_{\textcolor{black}{k}}) - \mathcal{L}(\mathbf{x}_{\textcolor{black}{i}},\mathbf{x}_{\textcolor{black}{j}}) - \mathcal{L}(\mathbf{x}_{\textcolor{black}{j}},\mathbf{x}_{\textcolor{black}{k}}) + \mathcal{L}(\mathbf{x}_{\textcolor{black}{j}},\mathbf{x}_{\textcolor{black}{j}}) - \mathcal{L}(\mathbf{x}_{\textcolor{black}{j}},\mathbf{x}_{\textcolor{black}{i}}) - \mathcal{L}(\mathbf{x}_{\textcolor{black}{k}},\mathbf{x}_{\textcolor{black}{j}}) + \mathcal{L}(\mathbf{x}_{\textcolor{black}{j}},\mathbf{x}_{\textcolor{black}{j}})
= d(\mathbf{x}_{\textcolor{black}{i}},\mathbf{x}_{\textcolor{black}{j}}) + d(\mathbf{x}_{\textcolor{black}{j}},\mathbf{x}_{\textcolor{black}{k}})$ \; \; \; \; \; \; \; \; \; \; \; \; \; \; \; \; \; \; \; \; \; \; \; \; \; \; \; \; \; \; \; $\square$

\subsection{KDSUM: \textcolor{black}{Kernel} Dissimilarity \textcolor{black}{M}etric for Mixed-Type Data} 
The pairwise similarity between two observations $\mathbf{x}_{i}$ and $\mathbf{x}_{j}$ is \begin{eqnarray}\label{eq:pairwise_dis}
\psi(\mathbf{x}_{i}, \mathbf{x}_{j} \vert \boldsymbol{\lambda}) = \prod_{k=1}^{p_c} \frac{1}{\lambda_k} k\left( \frac{x_{i,k} - x_{j,k}}{\lambda_k}\right)  + \sum_{k=p_c+1}^{p_u} L(x_{i,k},x_{j,k},\lambda_k) + \sum_{k = \textcolor{black}{p_c+}p_u+1}^p l(x_{i,k},x_{j,k},\lambda_k).
\end{eqnarray}

\textcolor{black}{Using the positive, definite, symmetric kernel functions defined} in Section \ref{sec:kernel}, $\psi(\cdot)$ satisfies the similarity properties (S1)-(S4) and is a similarity function, thus we can \textcolor{black}{set} $\mathcal{L}(\cdot):=\psi(\cdot)$. \textcolor{black}{We demonstrate that the sum of two similarity functions satisfies the rules (S1)-(S4) of being a similarity function, which can be extended to any number of sums of similarity functions. The results are similar for the product of similarity functions. We also note that since our similarity functions are kernel functions, the sum and product of multiple kernel functions is also a kernel function (see Bishop, 2006, pp 296).}\\

\textcolor{black}{\noindent\textit{Lemma 4:} The sum of similarity functions are also similarity functions}

\textcolor{black}{\textit{Proof}:}
    \textcolor{black}{Let $\mathcal{L}^{(n)}(x_i,x_j) = \mathcal{L}_1^{(n)}(x_i,x_j) + \ldots + \mathcal{L}_n^{(n)}(x_1,x_2)$ be the sum of $(n)$ similarity functions. We show each of the properties (S1)-(S4) through induction or directly: \\
    (S1): For the base case, $\mathcal{L}^{(2)}(x_i,x_j) = \mathcal{L}_1^{(2)}(x_i,x_j) +  \mathcal{L}_2^{(2)}(x_i,x_j) = \mathcal{L}_1^{(2)}(x_j,x_i) + \mathcal{L}_2^{(2)}(x_j,x_i) = \mathcal{L}^{(2)}(x_j,x_i)$. Then, assuming  $\mathcal{L}^{(n)}(x_i,x_j) = \mathcal{L}_1^{(n)}(x_i,x_j) + \ldots +  \mathcal{L}_n^{(n)}(x_i,x_j) = \mathcal{L}_1^{(n)}(x_j,x_i) + \ldots + \mathcal{L}_n^{(n)}(x_j,x_i) = \mathcal{L}^{(n)}(x_j,x_i)$ is true, we have $\mathcal{L}^{(n+1)}(x_i, x_j) = \mathcal{L}_1^{(n+1)}(x_i,x_j) + \ldots + \mathcal{L}_n^{(n)}(x_i,x_j) + \mathcal{L}_{n+1}^{(n+1)}(x_i,x_j) = \mathcal{L}_1^{(n+1)}(x_j,x_i) + \ldots + \mathcal{L}_n^{(n)}(x_j,x_i) + \mathcal{L}_{n+1}^{(n+1)}(x_j,x_i) = \mathcal{L}^{(n+1)}(x_j, x_i)$ \\
    (S2): If $x_i = x_j$, then their self-similarity values are maximum, and the similarity values between them is also maximum:
    \begin{align*}
\mathcal{L}^{(n)}(x_i, x_i) &= \text{maximum self-similarity value for the sum of $n$ similarity functions} \\
\mathcal{L}^{(n)}(x_i, x_j) &= \text{similarity value between } x_i \text{ and } x_j \text{ or the sum of $n$ similarity functions}\\
\mathcal{L}{(n)}(x_j, x_j) &= \text{maximum self-similarity value for the sum of $n$ similarity functions}
    \end{align*}
Since both $\mathcal{L}^{(n)}(x_i, x_i)$ and $\mathcal{L}^{(n)}(x_j, x_j)$ are maximum values, and the only way for $\mathcal{L}^{(n)}(x_i, x_j)$ to be the same as both $\mathcal{L}^{(n)}(x_i, x_i)$ and $\mathcal{L}^{(n)}(x_j, x_j)$ is if $x_i = x_j$.
Conversely, if $x_i \neq x_j$, then their self-similarity values are not maximum, and the similarity value between them should be less than the self-similarity values:
    \begin{align*}
\mathcal{L}^{(n)}(x_i, x_i) &= \text{self-similarity value of } x_i \text{ for the sum of $n$ similarity functions (not maximum)} \\
\mathcal{L}^{(n)}(x_i, x_j) &= \text{similarity value between } x_i \text{ and } x_j \text{ for the sum of $n$ similarity functions}\\
\mathcal{L}^{(n)}(x_j, x_j) &= \text{self-similarity value of } x_j \text{ for the sum of $n$ similarity functions (not maximum)}
    \end{align*}
Since $\mathcal{L}^{(n)}(x_i, x_i) \neq \mathcal{L}^{(n)}(x_i, x_j)$, and $\mathcal{L}^{(n)}(x_j, x_j) \neq \mathcal{L}^{(n)}(x_i, x_j)$, the only way for $\mathcal{L}^{(n)}(x_i, x_j)$ to be the same as both $\mathcal{L}^{(n)}(x_i, x_i)$ and $\mathcal{L}^{(n)}(x_j, x_j)$ is if $x_i = x_j$. In both cases, the property holds. \\
    (S3): For the base case, since based on properties of similarity, $\mathcal{L}_1^{(2)}(x_i,x_i) \geq \mathcal{L}_1^{(2)}(x_i,x_j) \geq 0$ and $\mathcal{L}_2^{(2)}(x_i,x_i) \geq \mathcal{L}_2^{(2)}(x_i,x_j) \geq 0$, then $\mathcal{L}_1^{(2)}(x_i,x_i) + \mathcal{L}_2^{(2)}(x_i,x_i) \geq \mathcal{L}_1^{(2)}(x_i,x_j) + \mathcal{L}_2^{(2)}(x_i,x_j) \geq 0 + 0$, and thus $\mathcal{L}^{(2)}(x_i,x_i) \geq \mathcal{L}^{(2)}(x_i,x_j) \geq 0$. \\
    Now, assuming $\mathcal{L}^{(n)}(x_i,x_i) \geq \mathcal{L}^{(n)}(x_i,x_j) \geq 0$, we note $\mathcal{L}_k^{(n+1)}(x_i,x_i) \geq \mathcal{L}_k^{(n+1)}(x_i,x_j) \geq 0$ for all $k \in \{1,2,\ldots,n+1\}$. Adding these $k$ sets of inequalities, and using our inductive hypothesis $\mathcal{L}^{(n+1)}(x_i,x_i) = \mathcal{L}_1^{(n+1)}(x_i,x_i) + \ldots + \mathcal{L}_n^{(n+1)}(x_i,x_i) + \mathcal{L}_{n+1}^{(n+1)}(x_i, x_i) \geq \mathcal{L}_1^{(n+1)}(x_i,x_j) + \ldots + \mathcal{L}_n^{(n+1)}(x_i,x_j) + \mathcal{L}_{n+1}^{(n+1)}(x_i, x_j) = \mathcal{L}^{(n+1)}(x_i,x_j) $ and  $\mathcal{L}^{(n+1)}(x_i,x_j) = \mathcal{L}_1^{(n+1)}(x_i,x_j) + \ldots + \mathcal{L}_n^{(n+1)}(x_i,x_j) + \mathcal{L}_{n+1}^{(n+1)}(x_i, x_j) \geq (n+1)\times 0$. \\
    (S4): For the base case, since based on properties of similarity, $\mathcal{L}_1^{(2)}(x_i,x_j) + \mathcal{L}_1^{(2)}(x_j,x_k) \leq \mathcal{L}_1^{(2)}(x_i,x_k) + \mathcal{L}_1^{(2)}(x_j,x_j)$, and $\mathcal{L}_2^{(2)}(x_i,x_j) + \mathcal{L}_2^{(2)}(x_j,x_k) \leq \mathcal{L}_2^{(2)}(x_i,x_k) + \mathcal{L}_2^{(2)}(x_j,x_j)$, then $\mathcal{L}_1^{(2)}(x_i,x_j) + \mathcal{L}_1^{(2)}(x_j,x_k) + \mathcal{L}_2^{(2)}(x_i,x_j) + \mathcal{L}_2^{(2)}(x_j,x_k) \leq \mathcal{L}_1^{(2)}(x_i,x_k) + \mathcal{L}_1^{(2)}(x_j,x_j) + \mathcal{L}_2^{(2)}(x_i,x_k) + \mathcal{L}_2^{(2)}(x_j,x_j) $, which implies $\mathcal{L}^{(2)}(x_i,x_j) + \mathcal{L}^{(2)}(x_j,x_k) \leq \mathcal{L}^{(2)}(x_i,x_k) + \mathcal{L}^{(2)}(x_j,x_j)$. \\
    Now, assuming $\sum_{N=1}^n\mathcal{L}_N^{(n)}(x_i,x_j) + \sum_{N=1}^n\mathcal{L}_N^{(n)}(x_j,x_k) \leq \sum_{N=1}^n\mathcal{L}_N^{(n)}(x_i,x_k) + \sum_{N=1}^n\mathcal{L}_N^{(n)}(x_j,x_j)$, then 
    \begin{eqnarray*}
    \mathcal{L}^{(n+1)}(x_i,x_j) + \mathcal{L}^{(n+1)}(x_j,x_k)  \\ = \sum_{N=1}^n\mathcal{L}_N^{(n+1)}(x_i,x_j) +  \mathcal{L}_{n+1}^{(n+1)}(x_i,x_j) + \sum_{N=1}^n\mathcal{L}_N^{(n+1)}(x_j,x_k) + \mathcal{L}_{n+1}^{(n+1)}(x_j,x_k) \\ \leq 
    \sum_{N=1}^n\mathcal{L}_N^{(n+1)}(x_i,x_k) + \mathcal{L}_{n+1}^{(n+1)}(x_i,x_k) + \sum_{N=1}^n\mathcal{L}_N^{(n+1)}(x_j,x_j) + \mathcal{L}_{n+1}^{(n+1)}(x_j,x_j) \\ = \mathcal{L}^{(n+1)}(x_i,x_k) + \mathcal{L}^{(n+1)}(x_j,x_j), 
    \end{eqnarray*}
    which holds true by the inductive hypothesis \\
    $\sum_{N=1}^n\mathcal{L}_N^{(n+1)}(x_i,x_j) + \sum_{N=1}^n\mathcal{L}_N^{(n+1)}(x_j,x_k) + \leq \sum_{N=1}^n\mathcal{L}_N^{(n+1)}(x_i,x_k) + \sum_{N=1}^n\mathcal{L}_N^{(n+1)}(x_j,x_j)$ and since $\mathcal{L}_{n+1}^{(n+1)}(x_i,x_j) + \mathcal{L}_{n+1}^{(n+1)}(x_j,x_k) \leq  \mathcal{L}_{n+1}^{(n+1)}(x_i,x_k) + \mathcal{L}_{n+1}^{(n+1)}(x_j,x_j)$ } \;  \; \; \; \; \; \; \; \; \; \; \; \;  \; \; \; \; \; \; \; \textcolor{black}{$\square$} 

Combining the similarity properties (S1)-(S4) and adapting the kernel distance described by Phillips and Venkatasubramanian (2011) to the multivariate setting, we define the kernel distance summation (KDSUM) metric between any two data points $\mathbf{x}_i$, $\mathbf{x}_j$ of the dataset $X$ as
\begin{equation}\label{eq:distfin}
    d(\mathbf{x}_i,\mathbf{x}_j\vert \boldsymbol{\lambda}) = \psi(\mathbf{x}_i,\mathbf{x}_i\vert \boldsymbol{\lambda}) + \psi(\mathbf{x}_j,\mathbf{x}_j\vert \boldsymbol{\lambda}) - 2\psi(\mathbf{x}_i,\mathbf{x}_j\vert \boldsymbol{\lambda}).
\end{equation}
\textcolor{black}{where $d(\mathbf{x}_i,\mathbf{x}_j\vert \boldsymbol{\lambda})=2\left(\psi(\mathbf{x}_i,\mathbf{x}_i\vert \boldsymbol{\lambda})-\psi(\mathbf{x}_i,\mathbf{x}_j\vert \boldsymbol{\lambda})\right)$ as $\psi(\mathbf{x}_i,\mathbf{x}_j\vert \boldsymbol{\lambda})$ is symmetric. Investigating the KDSUM metric asymptotics reveals that the bandwidth selection methodology described above is a shrinkage methodology between maximized dissimilarity and a fixed quantity of uniform dissimilarity between all points. To see KDSUM as a shrinkage method, consider that when $\mathbf{x_i}=\mathbf{x_j}$, then $\psi(\mathbf{x}_{i}, \mathbf{x}_{j} \vert \boldsymbol{\lambda})$ is maximized and $d(\mathbf{x}_{i}, \mathbf{x}_{j})=0$ at any value of $\boldsymbol{\lambda}$. Consider for continuous kernel types that:}
\begin{eqnarray*}
\textcolor{black}{\lim_{\lambda_k \to 0} \frac{1}{\lambda_k} k\left( \frac{x_{i,k} - x_{j,k}}{\lambda_k}\right)=\left\{\begin{matrix}
\infty, & x_{i,k}=x_{j,k},  \\ 
0, & x_{i,k}\neq x_{j,k},
\end{matrix}\right.} \\
\textcolor{black}{\lim_{\lambda_k \to \infty} \frac{1}{\lambda_k} k\left( \frac{x_{i,k} - x_{j,k}}{\lambda_k}\right)=0,\: \forall\: x_{i,k}, x_{j,k}.}
\end{eqnarray*}
\textcolor{black}{where $x_{i,k},x_{i,k}\in(-\infty,\infty)$. For the Aitken unordered kernel in Equation (\ref{eq:unordered}) with bandwidth support $\lambda_k=\left[0,1\right]$, then $L(x_{i,k},x_{j,k},0)=\mathds{1}(x_{i,k} = x_{j,k})$ and $L\left(x_{i,k},x_{j,k},1\right)=1$. For the Wang \& Van Ryzin ordered kernel in Equation (\ref{eq:ordered}) with bandwidth support $\lambda_k \in [0,1]$, then $l(x_{i,k},x_{j,k},0)=\textcolor{black}{\mathds{1}}(x_{i,k} = x_{j,k})$ and $l(x_{i,k},x_{j,k},1)=0$. Thus, the asymptotics for $d(\mathbf{x}_i,\mathbf{x}_j\vert \boldsymbol{\lambda})$ when $x_{i,k}\neq x_{j,k},\:\forall \: k$ are:}
\begin{eqnarray*}
\textcolor{black}{\lim_{\boldsymbol{\lambda}\to\mathbf{0}}d(\mathbf{x}_i,\mathbf{x}_j\vert \boldsymbol{\lambda})= 2\left(\prod_{k=1}^{p_c} \infty  + \sum_{k=p_c+1}^{p_u} 1 + \sum_{k = \textcolor{black}{p_c+}p_u+1}^p 1-\prod_{k=1}^{p_c} 0  + \sum_{k=p_c+1}^{p_u} 0 + \sum_{k = \textcolor{black}{p_c+}p_u+1}^p 0\right)=\infty,}
\end{eqnarray*}
\textcolor{black}{where the distance is clearly maximized. Alternatively, when all bandwidths are large \\ $\mathbf{\lambda}_\infty=\left(\infty,\infty,\ldots,\infty, \frac{g_1-1}{g_2},\frac{g_1-1}{g_2},\ldots,\frac{g_{p_u}-1}{g_{p_u}},1,1,\ldots,1\right)$, we find that}
\begin{eqnarray*}
\textcolor{black}{\lim_{\boldsymbol{\lambda}\to\mathbf{\lambda}_\infty}d(\mathbf{x}_i,\mathbf{x}_j\vert \boldsymbol{\lambda})= 2\left(\prod_{k=1}^{p_c} 0 + \sum_{k=p_c+1}^{p_u} 0 + \sum_{k = \textcolor{black}{p_c+}p_u+1}^p 0-\prod_{k=1}^{p_c} 0  + \sum_{k=p_c+1}^{p_u} 0 + \sum_{k = \textcolor{black}{p_c+}p_u+1}^p 0\right)=0,}
\end{eqnarray*}
\textcolor{black}{where all distances are equal to zero. Thus, this methodology is a shrinkage method between maximized differences and zero difference, depending on the choice of bandwidth. To select bandwidths, we employ a maximized similarity cross-validation (MSCV) approach similar to maximum likelihood cross-validation (MLCV) (Stone, 1974; Geisser, 1975). The main difference is that we replace the leave-one-out likelihood function $\mathcal{L}_{(-1)}(\mathbf{x}_i)$ with the leave-one-out similarity function $\psi_{(-1)}(\mathbf{x}_{i} \vert \boldsymbol{\lambda})=\frac{1}{n-1}\sum_{j=1,j\neq i}^n \psi(\mathbf{x}_i,\mathbf{x}_j\vert \boldsymbol{\lambda})$ yielding $CV(\boldsymbol{\lambda}) = \sum_{i=1}^n\ln\left(\psi_{(-i)}
(\mathbf{x}_{i} \vert \boldsymbol{\lambda})\right).$ The advantage of this methodology is that we are minimizing the dissimilarity between data points for continuous and categorical variables simultaneously, which allows for the clustering of similar points, while smoothing out irrelevant variables not important for similarity or clustering. \\} 

\textcolor{black}{The algorithm to calculate the KDSUM metric is:}
\begin{algorithm}
    \caption{KDSUM}
    \begin{algorithmic}[1]
	\State Given a dataset $X$, reorder as \textcolor{black}{ continuous, unordered categorical, then ordered categorical, and ensure that the variables are casted accordingly.}
    \State Select symmetric kernel functions $K, L, l$ from \textcolor{black}{Equations (\ref{eq:gauss})-(\ref{eq:ordered}), or any symmetric kernels of choice.}
    \State Calculate optimal bandwidths for each $p_i$ using cross-validation procedure \textcolor{black}{outlined in Section \ref{sec:kernel}. 
    \State i. Define $\psi_{-i}\textbf{x}_i,\textbf{x}_j,\bm{\lambda})$, which is the updated leave-one-out pairwise similarity defined in Equation (\ref{eq:pairwise_dis}) replacing $\mathcal{L}_{-i}$ in Equation (\ref{eq:cvml}), ensuring the kernels selected in Step 2 are consistent. 
    \Statex ii. Optimize the function in Equation (\ref{eq:cvml}) to obtain the optimal bandwidths $\bm{\lambda}$. To do a leave-one-out on $\psi_{-i}(\cdot)$, remove one observation at a time and use the average to obtain $\bm{\hat{\lambda}}$. Optimization is achieved through quadratic optimization subject to the bandwidth range constraints.
    \Statex iii. The obtained $\bm{\hat{\lambda}}$ is considered the optimal bandwidth, and is the vector of minimal parameters to maximize the separation of observations for each variable.}
    \State Calculate the pairwise distance between all observations $\mathbf{x}_i$ and $\mathbf{x}_j$ using Equation (\ref{eq:distfin}) and the selected kernels in Step 2, \textcolor{black}{with optimal bandwidth from Step 3, to obtain the $n \times n$ dissimilarity matrix.}
    \end{algorithmic}
\end{algorithm}

Consider the following \textcolor{black}{toy example of a} simulated mixed-type data matrix \textcolor{black}{to illustrate how distances are smoothed through bandwidth selection}: \begin{singlespace}\[X = \begin{blockarray}{cccc} & p_{c_1} & p_{u_1} & p_{o_1} \\ \begin{block}{c(ccc)} \mathbf{x}_1 & 1.5 & 1 & 3 \\ \mathbf{x}_2 & 1.5 & 1 & 3 \\ \mathbf{x}_3 & 1.5 & 0 & 0 \\ \mathbf{x}_4 & 0 & 1 & 0 \\ \mathbf{x}_5 & 0 & 0 & 3 \\ \end{block} \end{blockarray}\:.\]\end{singlespace} \textcolor{black}{The columns are the continuous $p_{c_1}$, unordered $p_{u_1}$, and order $p_{o_1}$ variables. } The observered vectors $\mathbf{x}_1$ \textcolor{black}{and} $\mathbf{x}_2$ \textcolor{black}{are identical and thus the distance between them is} assigned 0\textcolor{black}{. The vectors} $\mathbf{x}_3$, $\mathbf{x}_4$, and $\mathbf{x}_5$ each contain one value in common with both $\mathbf{x}_1$ and $\mathbf{x}_2$ but the rest of the observations are 0, and the variables in common between $\mathbf{x}_3$, $\mathbf{x}_4$, and $\mathbf{x}_5$ and $\mathbf{x}_1$ and $\mathbf{x}_2$ are different for each vector. \\

\textcolor{black}{Consider three cases for} the bandwidth vector\textcolor{black}{s: (1) A specified small bandwidth vector } $\boldsymbol{\lambda}_1 = [0.01, 0, 0]$\textcolor{black}{, (2) a specified large bandwidth vector} $\boldsymbol{\lambda}_2 = [10, 1, 1]$\textcolor{black}{, and bandwidths selected} using maximum \textcolor{black}{similarity} cross\textcolor{black}{-}validation  \textcolor{black}{$\boldsymbol{\lambda}_3 = [1.027, 0.591, 4.94\times10^{-32}]$.} We observe that the variable $p_{u_1}$ has nearly reached its upper bounds and will contribute little to the overall distance based on the data, while $p_{c_1}$ and $p_{o_1}$ will contribute more heavily. The results are shown in case 3. \\

\begin{singlespace}
case 1: $d(X \; | \; \boldsymbol{\lambda}_1)$
\[\begin{blockarray}{cccccc}
    & \mathbf{x}_1 & \mathbf{x}_2 & \mathbf{x}_3 & \mathbf{x}_4 & \mathbf{x}_5 \\
    \begin{block}{c(ccccc)}
    \mathbf{x}_1 & 0 & 0 & 4.000 & 81.788 & 81.788 \\
    \mathbf{x}_2 &  & 0 & 4.000 & 81.788 & 81.788 \\
    \mathbf{x}_3 &  &  & 0 & 81.788 & 81.788 \\
    \mathbf{x}_4 &  &  &  & 0 & 4.000 \\
    \mathbf{x}_5 &  &  &  &  & 0 \\
  \end{block}
  \end{blockarray}
\]
case 2: $d(X \; | \; \boldsymbol{\lambda}_2)$
\[\begin{blockarray}{cccccc}
    & \mathbf{x}_1 & \mathbf{x}_2 & \mathbf{x}_3 & \mathbf{x}_4 & \mathbf{x}_5 \\
    \begin{block}{c(ccccc)}
    \mathbf{x}_1 & 0 & 0 & 0 & 0.001 & 0.001 \\
    \mathbf{x}_2 & & 0 & 0 & 0.001 & 0.001 \\
    \mathbf{x}_3 & & & 0 & 0.001 & 0.001 \\
    \mathbf{x}_4 & & & & 0 & 0 \\
    \mathbf{x}_5 & & & & & 0 \\
  \end{block}
  \end{blockarray}
\]
case 3: $d(X \; | \; \boldsymbol{\lambda}_3)$
\textcolor{black}{\[\begin{blockarray}{cccccc}
    & \mathbf{x}_1 & \mathbf{x}_2 & \mathbf{x}_3 & \mathbf{x}_4 & \mathbf{x}_5 \\
    \begin{block}{c(ccccc)}
    \mathbf{x}_1 & 0 & 0 & 2.819& 2.510 & 1.328 \\
    \mathbf{x}_2 & & 0 & 2.819 & 2.510 & 1.328 \\
    \mathbf{x}_3 & & & 0 & 1.328 & 2.510 \\
    \mathbf{x}_4 & & & & 0 & 2.819 \\
    \mathbf{x}_5 & & & & & 0 \\
  \end{block}
  \end{blockarray}
\]}
\end{singlespace}

\section{Study Description\textcolor{black}{s and Results}} \label{sec:data} 
To evaluate the performance of the KDSUM metric in comparison to established metrics for mixed-type data distance-based clustering, we analyzed simulated and real datasets of continuous, categorical, and mixed-type attributes using agglomerative hierarchical clustering techniques. We establish the performance of the KDSUM metric relative to existing metrics for handling mixed-type data and demonstrate the potential of the KDSUM metric to enhance CA. By comparing the KDSUM metric to these advanced clustering techniques, we demonstrate the flexibility of the KDSUM metric for usage in \textcolor{black}{the} clustering of mixed datasets. \\

\subsection{Clustering algorithms}\label{sec:cluster}
Mixed-type approaches offer a solution to the challenge of clustering datasets that contain both continuous and categorical variables. One approach involves selecting a distance metric that can handle both types of variables, and then clustering the data using methods that depend on the distance function.  \\

\subsubsection{\textcolor{black}{Clustering with Distance Metrics}}\label{subsec:clustmethodology}
A kernel distance metric can be utilized in any clustering algorithm that accepts a dissimilarity metric. Additionally, this metric can be adapted to centroid, medoid, or prototype-based methods. To test the KDSUM metric for clustering, we follow previous literature that uses agglomerative hierarchical clustering algorithms designed to cluster based on dissimilarity metrics (see, for example, Day \& Edelsbrunner, 1984; Murtagh \& Contreras, 2012; Bouguettaya et al., 2015; Sasirekha \& Baby, 2013; Nielsen, 2016). 

Single-linkage calculates the distance between two clusters as the shortest distance between any two points in the two clusters. Similarly, Complete-linkage (e.g., Macnaughton-Smith, 1965) calculates the distance between two clusters as the maximum distance between any two points in the two clusters. Average-linkage (e.g., Lance \& Williams, 1967), on the other hand, considers the average distance between all pairs of points in the two clusters. Ward's method (Ward, 1963) seeks to minimize the total variance within each cluster as the criterion for merging clusters. Median linkage employs the median distance between all pairs of points in the two clusters, while centroid linkage (e.g., Sokal \& Michener, 1958) relies on the distance between the centroids of the two clusters. \\

\textcolor{black}{For the simulated and empirical application that follow, the KDSUM metric with Gaussian, Aitken, and  Wang \& van Ryzin kernels is used in a modified $k$-means clustering algorithm and competing hierarchical clustering methods, including single-, average-, and complete-linkages along with Ward's method. We report only the method with the highest CA and ARI is reported when utilizing the KDSUM to compare to competing methods.} \\

\subsubsection{\textcolor{black}{Clustering} Evaluation} 
When evaluating and comparing the effectiveness and accuracy of clustering and classification techniques, we use the two commonly used metrics of CA and the ARI. \textcolor{black}{We chose to evaluate our clustering results using both ARI and CA to provide a comprehensive and well-rounded assessment of our proposed clustering algorithm. While CA offers a straightforward measure of correct assignments, the ARI considers both pairwise agreements and disagreements, normalized for chance. This enables us to better understand the clustering performance in scenarios where clusters may be of varying sizes and complexities, where CA results may become more difficult to interpret.} \\

The ARI is a statistic that quantifies the similarity between the true classification of the data and the classification obtained by a given method (Rand, 1971). The ARI is defined as
$$ARI = \frac{ \sum_{ij} { {n_{ij}}\choose{2} } - [ \sum_{i} { {a_{i}}\choose{2} } \sum_{j} { {b_{j}}\choose{2} } ] / { {n}\choose{2} } } { \frac{1}{2} [ \sum_{i} { a_{i}\choose{2} } + \sum_{j} { {b_{j}}\choose{2} } ] - [ \sum_{i} { {a_{i}}\choose{2} } \sum_{j} { {b_{j}}\choose{2} } ] / { {n}\choose{2} } }, $$
where $n_{ij}$ is the diagonal sum of the clustering contingency table, and $a_i$, $b_j$ correspond to the row sums and column sums of the contingency table, respectively. The contingency table is a visual depiction that summarizes agreeance and disagreeance between the true class labels and the classification class labels. The index considers the number of pairs of data points that are labelled identically in both sets and labelled differently in both sets. The ARI then adjusts for a chance agreement based on the expected agreement between the two sets under a null model. The resulting ARI value ranges from 0 to 1, where 0 indicates complete randomness and 1 indicates perfect agreement in classification. \\

CA is also used to measure the percentage of data points correctly assigned to their corresponding clusters. It is calculated by comparing the true classification labels with those generated by the clustering algorithm, defined as $$CA(y,\hat{y}) = \frac{\sum_{i=1}^n\textcolor{black}{\mathds{1}}(\hat{y}_i = y_i)}{n},$$ where the indicator function $\textcolor{black}{\mathds{1}}(\textcolor{black}{\hat{y}_i = y_i}) = 1$ if the class label \textcolor{black}{$\hat{y}_i = y_i$}, and 0 otherwise. The CA ranges from 0 to 1, where 0 indicates that none of the data points are assigned to the correct clusters, and 1 indicates that all data points are assigned to the correct clusters.\\

\subsection{Simulat\textcolor{black}{ion Studies}}

\textcolor{black}{This section describes the simulated data used to investigate the performance of the KDSUM metric for clustering using the methodologies described in Section \ref{sec:cluster}. We analyze Monte Carlo simulations for each clustering algorithm using the mixed-type metrics on continuous-only, categorical-only, and mixed-type simulated datasets.}

\subsubsection{\textcolor{black}{Continuous data}}
\begin{figure}[H]
    \centering
    \includegraphics[width = \textwidth]{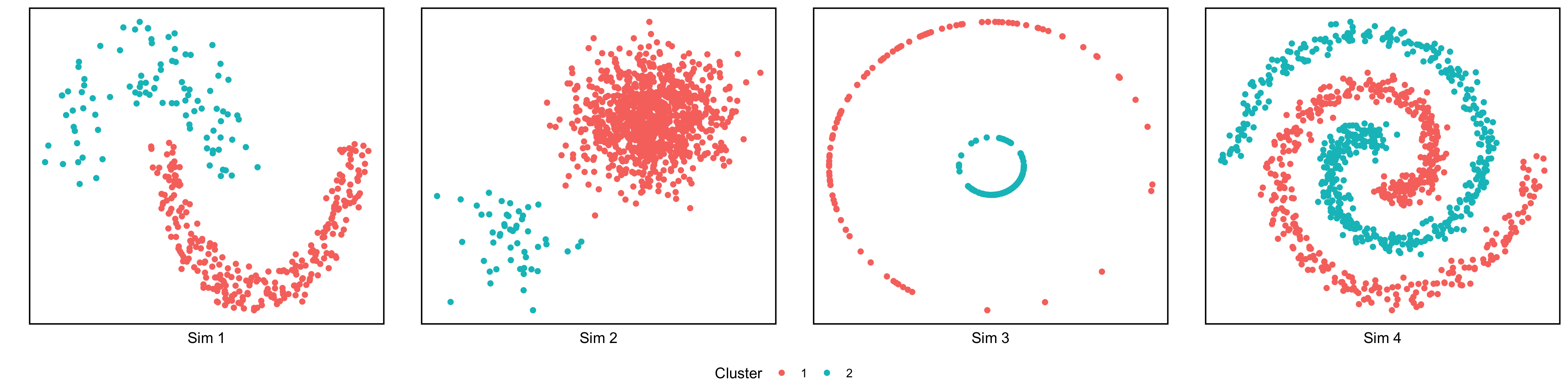}
    \caption{Variable distribution with respect to cluster assignment for four continuous simulated datasets. From left to right: Sim 1, Sim 2, Sim 3, Sim 4.}
    \label{fig:cont4viz}
\end{figure}

The first four continuous \textcolor{black}{simulated} datasets \textcolor{black}{were adapted from Morbieu (2018)} to evaluate the ability of KDSUM to effectively handle data that exhibit \textcolor{black}{linear and nonlinear clustering patterns}. In all instances, the simulated data comprised two variables and two known classes. Specifically, the first \textcolor{black}{dataset (Sim 1)} consisted of 373 observations simulated \textcolor{black}{as in Figure \ref{fig:cont4viz}, where each Monte Carlo iteration allowed each observation to shift in in four directions. The shift of each observation was drawn from a uniform distribution between $\pm 0.5$}. The second \textcolor{black}{dataset (Sim 2)} contained 2050 observations that were simulated using a well-defined large cluster of 2000 observations with low variance, and a small cluster of 50 observations with high variance. The third \textcolor{black}{dataset (Sim 3)} consisted of 200 observations that were simulated with one dense spherical cluster that was contained inside a sparse spherical cluster, with both clusters having equal observations. Lastly, the fourth \textcolor{black}{dataset (Sim 4)} consisted of two equally-sized clusters that were spiralled within each other. A visualization of the four simulated continuous datasets is presented in Figure \ref{fig:cont4viz}.

\subsubsection{\textcolor{black}{Categorical data}}
\begin{figure}[H]
    \centering
    \includegraphics[width = \textwidth]{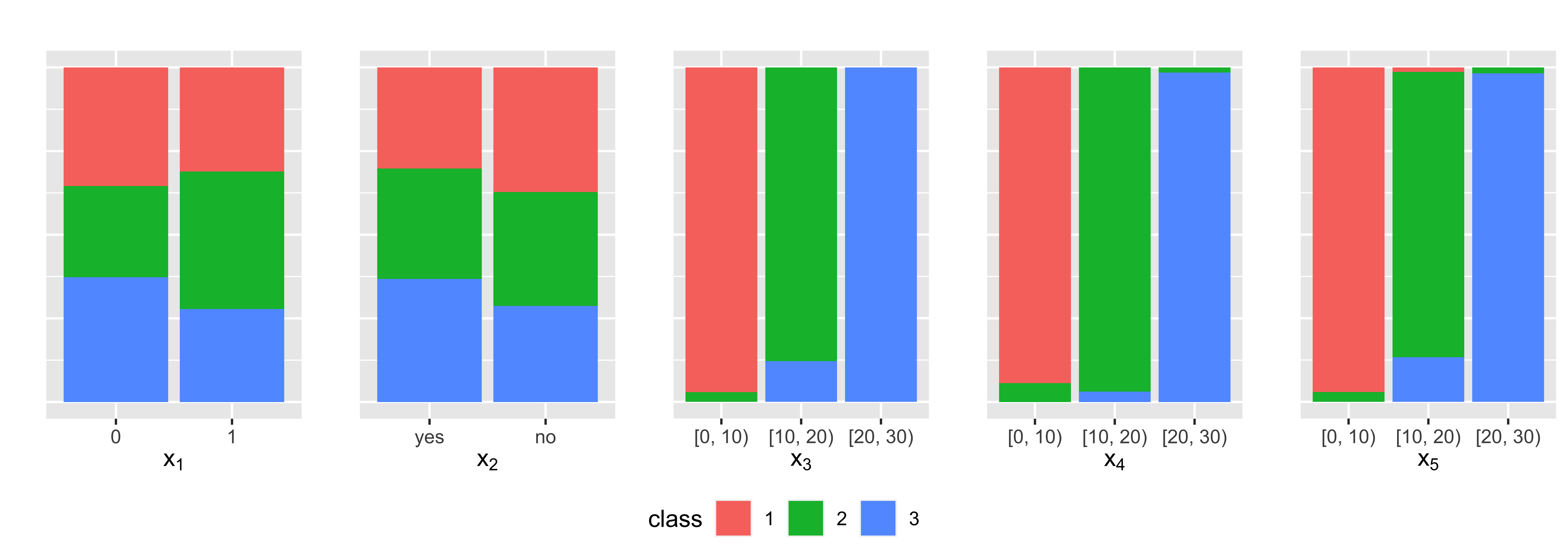}
    \caption{Variable distribution with respect to cluster assignment for Sim 5. $X_1$ and $X_2$ represent the binary noise variables, and $X_3$, $X_4$ and $X_5$ are the meaningful categorical variables, grouped in a 10 unit interval for ease of interpretation.}
    \label{fig:sim5viz}
\end{figure}

\textcolor{black}{A} categorical\textcolor{black}{-only dataset (Sim 5) consisted of $200$ observations, with $5$ unordered categorical variables and $3$ clusters. Two of the variables were random binary noise variables, and the three unordered categorical variables were randomly selected integers in the \textcolor{black}{interval} $\textcolor{black}{[}0\textcolor{black}{,}30\textcolor{black}{)}$, and the first cluster consisted only of values \textcolor{black}{in} $\textcolor{black}{[}0\textcolor{black}{,}10\textcolor{black}{)}$ for each of the three variables, while classes two and three consisted of values \textcolor{black}{in} $\textcolor{black}{[}10\textcolor{black}{,}20\textcolor{black}{)}$ and $\textcolor{black}{[}20\textcolor{black}{,}30\textcolor{black}{)}$, respectively.} Figure \ref{fig:sim5viz} \textcolor{black}{shows a single simulation from this dataset, which shows} that there \textcolor{black}{is} some overlap in \textcolor{black}{the} cluster assignment\textcolor{black}{s}.

\subsubsection{\textcolor{black}{Mixed-type data}}

\textcolor{black}{Sim 6 was constructed with $373$ observations and $5$ variables, where $2$ continuous variables followed the same distribution as Sim 1 with the same variation at each Monte Carlo iteration. The one unordered categorical variable, $X_3$, was binary and generated randomly as noise at each iteration. The two unordered categorical variables ranged from $0$ to $2$ for the first cluster and $3$ to $4$ for the second, randomly drawn at each iteration from a uniform distribution then rounded to an integer. Cluster 1 contained 97 observations, while cluster 2 contained 276. }

\subsubsection{\textcolor{black}{Results}}

\begin{figure}[htbp]
    \centering
    \includegraphics[width = \textwidth]{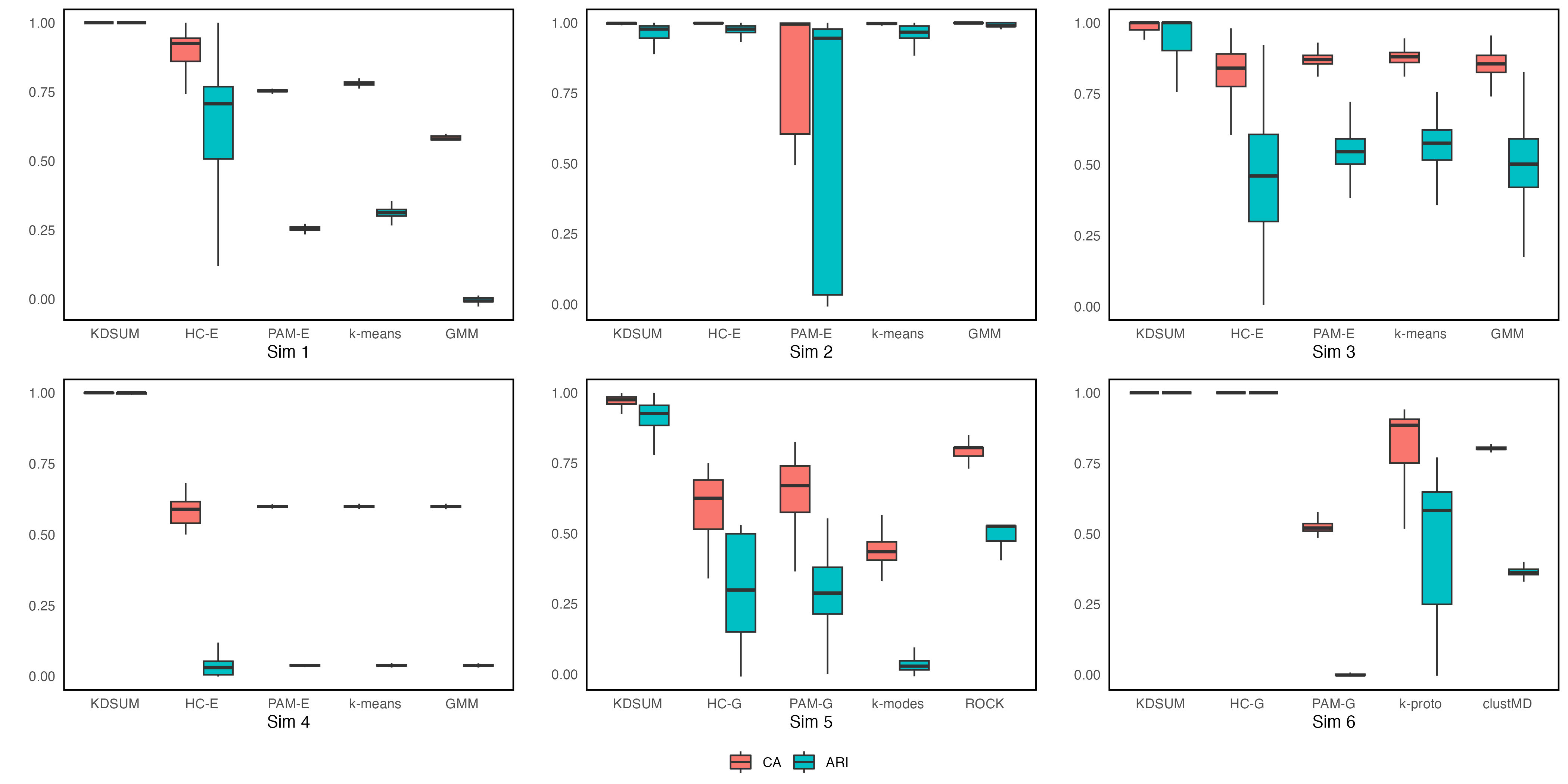}
    \caption{\textcolor{black}{Boxplots of ARI and CA for KDSUM with hierarchical clustering and average-linkage, compared against competing clustering algorithms for simulated continuous, categorical, and mixed-type data.}}
    \label{fig:sim1-6mcres}
\end{figure}

\textcolor{black}{For each of the six data generating processes described in the previous sections, we conducted 1000 Monte Carlo simulations and analyzed them using KDSUM with hierarchical clustering and average-linkage compared to the clustering algorithms in Section \ref{subsec:clustmethodology}. Distributions and average values of the CA and ARI for clustering Monte Carlo simulations are shown in Figure \ref{fig:sim1-6mcres} and Table \ref{tab:results_simulated}, respectively. For Sim 1, KDSUM hierarchical clustering outperformed each of the other methods and had nearly perfect CA. For Sim 2, GMM had the highest CA and ARI, slightly higher than KDSUM hierarchical clustering. This is likely due to the two clusters being circular with some overlap, where a linear mixture model is a correctly specified clustering algorithm for this datatype. For Sim 3 and Sim 4, we see that KDSUM is clustering well on these and demonstrates that the KDSUM metric is able to estimate non-linear datasets. Sim 5 and Sim 6 both contain large values of CA and ARI for KDSUM hierarchical clustering while in the presence of categorical and continuous noise variables. We note for Sim 6 that Gower's distance with hierarchical clustering uses Euclidean distance and a simple matching coefficient with a weight for distance measurement. Gower's distance tends to place more emphasis on the categorical variables as a result, and thus performs well since there is low overlap between categorical variables in terms of clustering. The bandwidths selected via MSCV for KDSUM were very small; thus the KDSUM metric aligned with Gower's distance in the metric calculation.}
\begin{table*}[htbp]
\centering
\caption{\textcolor{black}{Classification results on Monte Carlo simulated data. The KDSUM metric with agglomerative hierarchical clustering was compared to Partitioning around Medoids (PAM-E / PAM-G), Euclidean or Gower's distance with hierarchical clustering (HC-E / HC-G),  $k-$means, $k-$modes, $k-$prototypes ($k-$proto), Gaussian Mixture Model (GMM), ROCK, and clustMD. For hierarchical clustering, the method is reported in brackets. The average ARI and CA of 1000 Monte Carlo simulations is reported for each clustering algorithm.}}
{\color{black}\begin{tabular}{cccc|cccc}
\hline
\hline
Data & Model & CA & ARI & Data & Model & CA & ARI\\
\hline
\hline
Sim 1       & KDSUM (Average)  & 0.990 & 0.960 &    Sim 4 & KDSUM (Average) & 0.983 & 0.962 \\
(cont.)& HC-E (Average) & 0.880 & 0.604 & & HC-E (Average) & 0.580 & 0.033 \\
            & PAM-E & 0.752 & 0.253  &              (cont.) & PAM-E & 0.599 & 0.038 \\
            & $k-$means & 0.780 & 0.312 &           & $k-$means & 0.599 & 0.038 \\
            & GMM & 0.577 & 0.038 & & GMM & 0.592 & 0.035 \\ \hline
            
Sim 2       & KDSUM (Average)  & 0.989 & 0.929 & Sim 5      & KDSUM (Average)  & 0.938 & 0.860  \\ 
(cont.)     & HC-E (Average) & 0.998 & 0.974 & & HC-G (Average) & 0.594  &0.311 \\ 
            & PAM-E & 0.817 & 0.543 & (cat.) & PAM-G & 0.648 & 0.230  \\
            & $k-$means & 0.997 & 0.963 & & $k-$modes & 0.439  & 0.034  \\
            & GMM & 0.999 & 0.989 & & ROCK & 0.787 & 0.498 \\ \hline
Sim 3       & KDSUM (Average) & 0.913 & 0.811 &  Sim 6      & KDSUM (Average) & 1.000 & 1.000 \\
(cont.)           & HC-E (Average)  & 0.820 & 0.444 & & HC-G (Average) & 1.000 & 1.000 \\
            & PAM-E & 0.869 & 0.545 & (mix.) & PAM-G & 0.523 & 0.000 \\
            & $k-$means & 0.877 & 0.569 &  & $k-$proto & 0.828 & 0.459 \\
            &  GMM & 0.854  & 0.507 & & clustMD & 0.802 & 0.363 \\
\hline
\hline
\end{tabular}}
\label{tab:results_simulated}
\end{table*}

\subsection{\textcolor{black}{Bandwidth Grid Search}}

\textcolor{black}{In this section, we establish the effect of KDSUM bandwidth selection in clustering. The choice of bandwidths can affect the distance metric calculation, and we show that MSCV selects ideal bandwidths that are small for variables relevant to clustering and large for variables that are irrelevant. Clustering for this section was conducted using agglomerative hierarchical clustering with single-linkage. To examine bandwidth selection influence on clustering through CA and ARI, we analyze simulated data consisting of continuous-only, categorical-only, and mixed-type data. \\}

\subsubsection{\textcolor{black}{Continuous data}}
\begin{figure}[H]
    \centering
    \includegraphics[width = \textwidth]{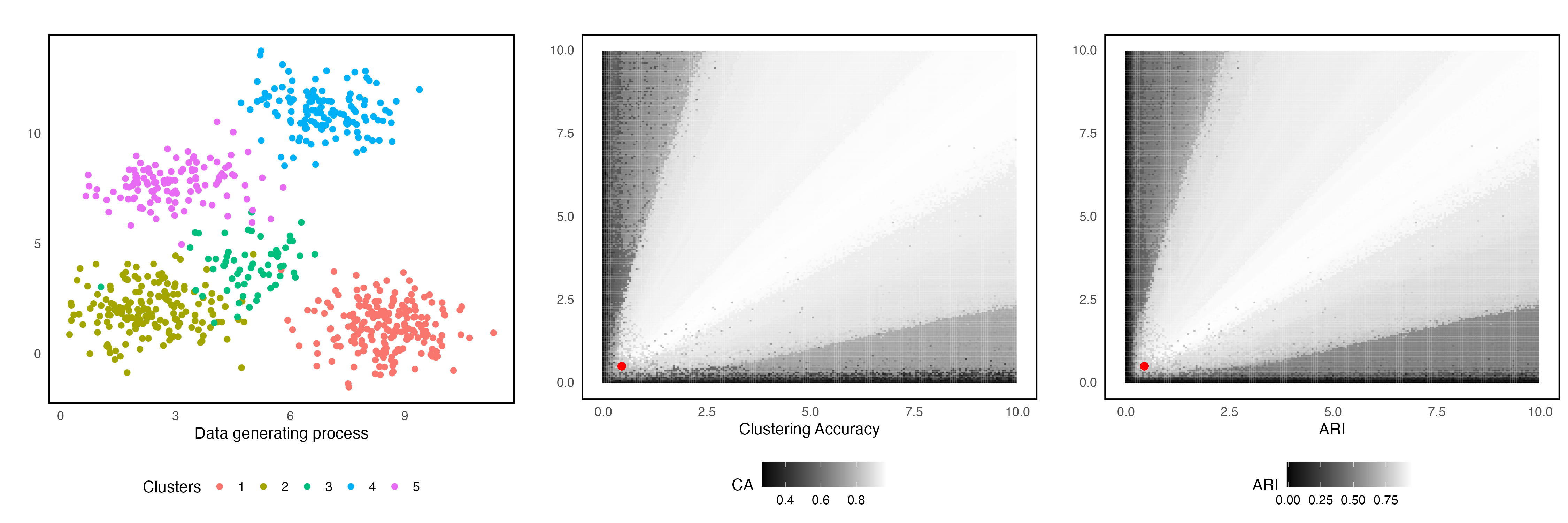}
    \caption{\textcolor{black}{The left panel is the data generating process of two continuous variables $X_1$ and $X_2$, with $k = 5$ clusters. The middle and right panels depict the CA and ARI, respectively, for the continuous bandwidth grid search, where increments of bandwidths were 0.05 in the range $[0,10]$ for both variables. The red dot on each panel indicates the optimal bandwidth selected via MSCV for the KDSUM metric.}}
    \label{fig:cont_gridsearch}
\end{figure}

\textcolor{black}{For the continuous-only data, we simulated two continuous variables and five distinct clusters. The cluster centres are selected at random from a two-dimensional uniform distribution spanning the interval $[0, 12]\times [0, 12]$, with each cluster populated through a multivariate normal distribution, with cluster sample sizes selected randomly from 50 to 200. A visual representation of the data is shown in the left panel of Figure \ref{fig:cont_gridsearch}. A systematic grid search was conducted, spanning bandwidth values from 0 to 10 and incremented in steps of 0.05 and results from clustering using the modified $k-$means for distance matrices algorithm, are shown in the right two panels of Figure \ref{fig:cont_gridsearch}. The MSCV selected bandwidth values are $(0.443, 0.483)$ for $X_1$ and $X_2$, respectively. We can see that the performance metrics remain high at elevated bandwidths; however, these two bandwidths must be balanced to allow KDSUM to accurately measure dissimilarity and improve underlying clustering algorithm accuracy. This plot demonstrates that MSCV preferentially selects the smallest bandwidths for significant variables.\\}

\subsubsection{\textcolor{black}{Categorical data}}
\begin{figure}[H]
    \centering
    \includegraphics[width = \textwidth]{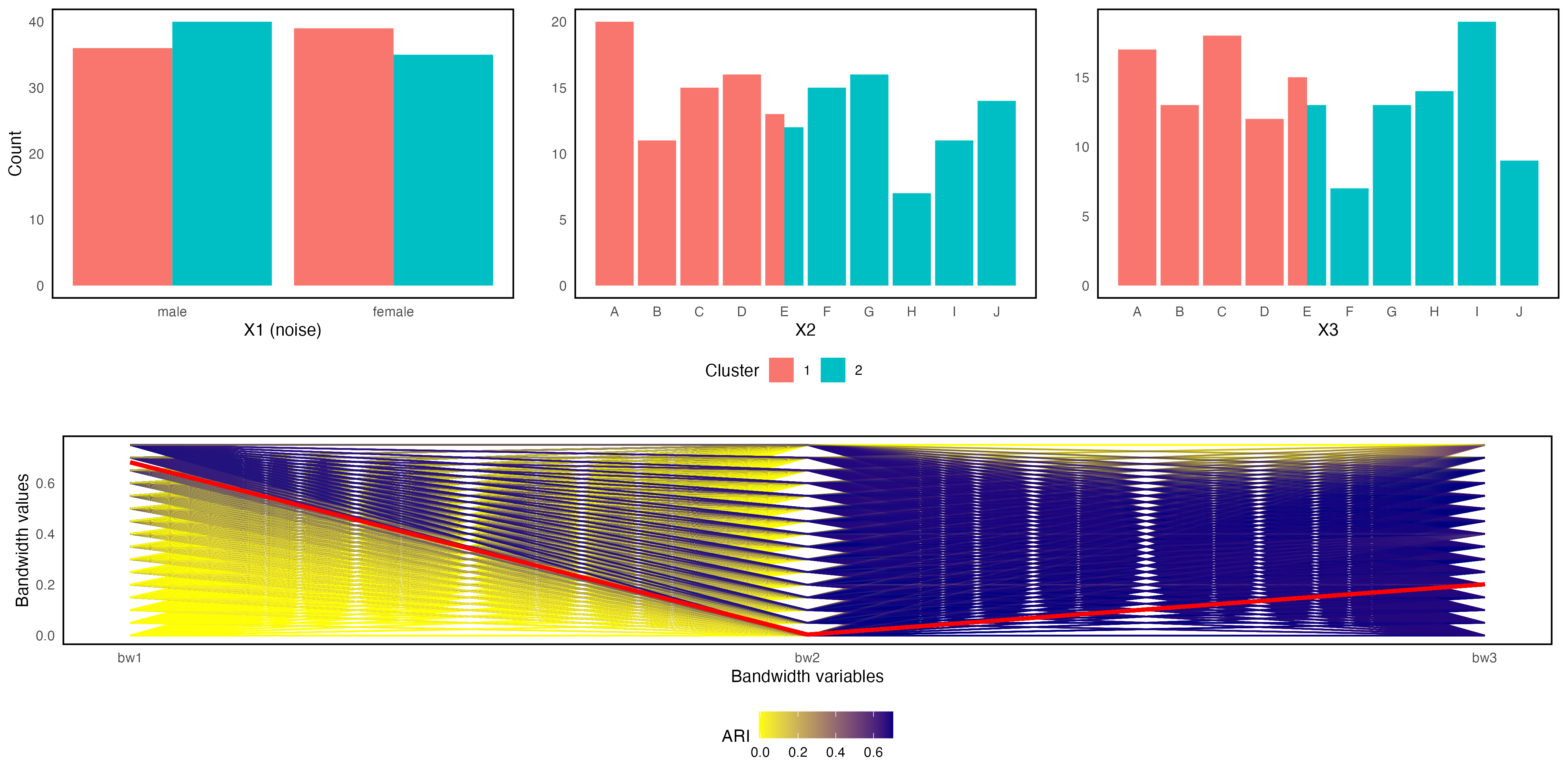}
    \caption{\textcolor{black}{The upper three panels are the data generating process with $k = 2$ clusters, with binary noise term ($X_1$), and two unordered categorical variables ($X_2, X_3$). The bottom plot is a parallel coordinates plot for the unordered categorical bandwidth grid search, where increments of bandwidths were 0.05 in the range $[0,0.75]$ for all variables, coloured by the ARI for each possible combination. The red lines indicate the optimal bandwidth determined using maximum-likelihood cross-validation with the KDSUM metric.}}
    \label{fig:cat_gridsearch}
\end{figure}

\textcolor{black}{Next, we investigate a simulated categorical-only dataset. The data is generated with three distinct clusters based on a binary noise term ($X_1$) and two categorical variables ($X_2$ \& $X_3$). The first cluster is generated with random integers 1 through 5, the second with random integers 5 through 10. Other than the noise term, the overlap in the ranges of the integers of each cluster allows for cluster overlap. The ratio of the three clusters is 1:1, respectively, with 75 observations per cluster. All variables are treated as unordered categorical. A visual representation of this data is shown in the top three panels of Figure \ref{fig:cat_gridsearch}. \\}

\textcolor{black}{The resultant grid search entails increments of $0.05$ for each of the three variables in the range $[0,\frac{3}{4}]$, resulting in $4,096$ permutations. The results, presented in the bottom panel of Figure \ref{fig:cat_gridsearch}, show that clustering performance is best when we use small bandwidths for relevant variables and large bandwidths for irrelevant variables. The noise variable is shown to provide poor performance (small ARI) when the bandwidth is small, and the optimal bandwidth is a large value that effectively smooths out the noise variable for distance calculations and thus becomes irrelevant to clustering. The important variables for clustering are the two categorical where MSCV selects small bandwidths and provides the highest ARI compared to any other set of bandwidths.}

\subsubsection{\textcolor{black}{Mixed-type data}}
\begin{figure}[H]
    \centering
    \includegraphics[width = \textwidth]{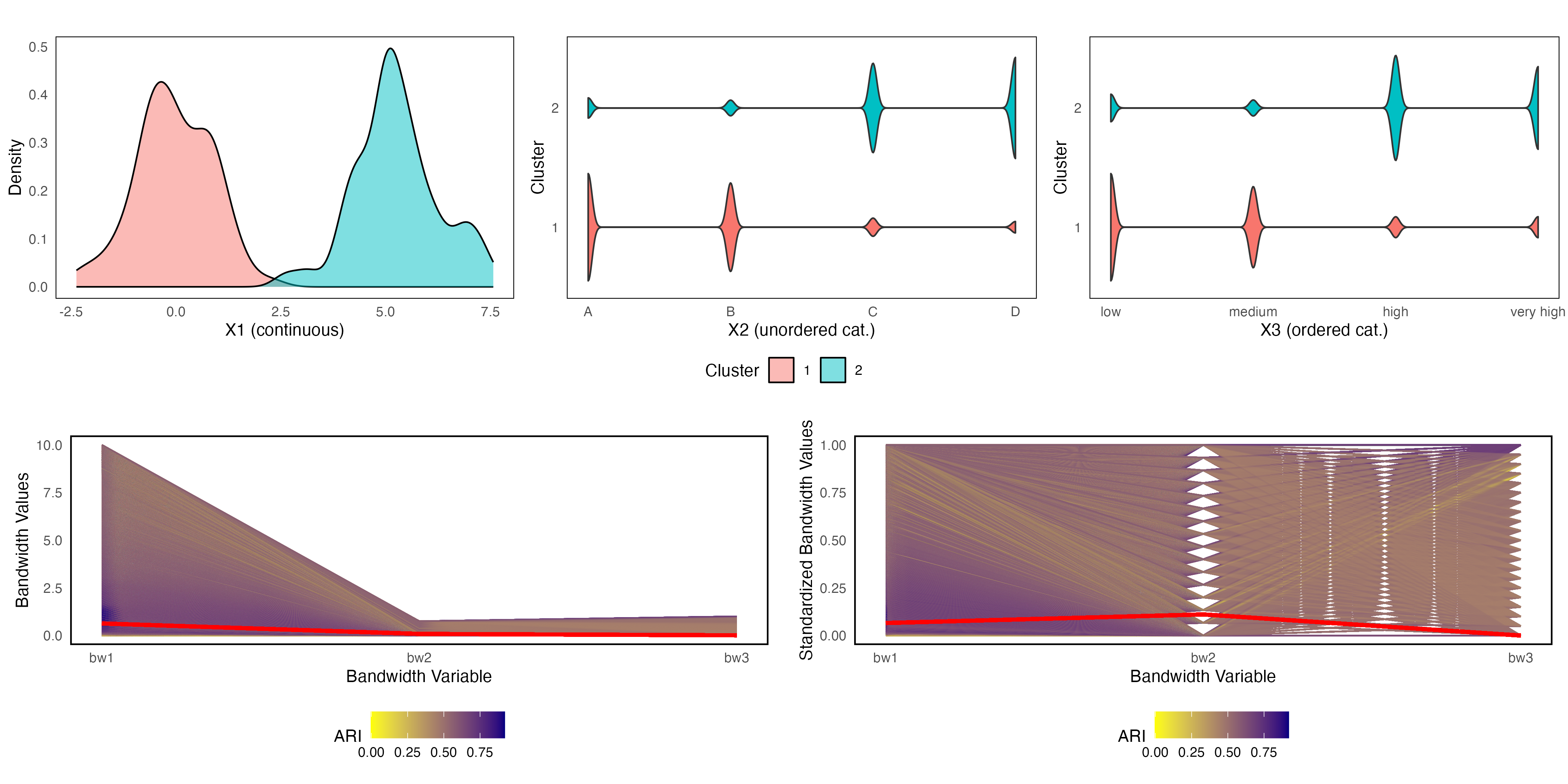}
    \caption{\textcolor{black}{The upper three panels are the data generating process with $k=2$ cluster, with one continuous $(X_1)$, unordered categorical ($X_2$), and one ordered ($X_3$) variable. Each of the two categorical variables has four levels. The parallel coordinates plots depicted below are the regular and standardized version of the bandwidth values for each variable, for ease of interpretation. The red lines indicate the regular and standardized version of the optimal bandwidth determined using maximum-likelihood cross-validation in conjunction with the KDSUM method.}}
    \label{fig:bwgrid_mix}
\end{figure}

\textcolor{black}{We extend this simulation to a mixed dataset that consists of a mixture of the previous two datasets; 200 observations consisting of one continuous variable generated from a normal mixture model, and two categorical variables (one unordered and one ordered) generated from a multinomial mixture model. The overlap between continuous and categorical variables is set to 5\% for one continuous and 35\% for two categorical variables. The 200 observations are portioned to a $40:60$ ratio among the two clusters. A visual representation of this data is shown in the top three panels of Figure \ref{fig:bwgrid_mix}.\\}

\textcolor{black}{The resultant grid search entails varied bandwidth assignments: the continuous variable's ($X_1$) bandwidth is incremented in steps of 0.05 over the interval $[0, 10]$, the unordered categorical variable $X_2$ range in $[0, \frac{3}{4}]$ at increments of 0.05, while the ordered categorical variable $X_3$ in $[0, 1]$, also in increments of 0.05, resulting in $67,536$ permutations. Clustering for this simulation was completed with the modified $k-$means algorithm for distance matrices, and the results are presented in the bottom two panels of Figure \ref{fig:bwgrid_mix}. The optimal set of bandwidths selected by MSCV shows that each variable is relevant to the clustering algorithm, where large bandwidths are shown to cause suboptimal ARI in clustering.}

\subsection{\textcolor{black}{Effect of Sample Size}}\label{subsec:MCSvcs}

\begin{figure}[H]
    \centering
    \includegraphics[width = \textwidth]{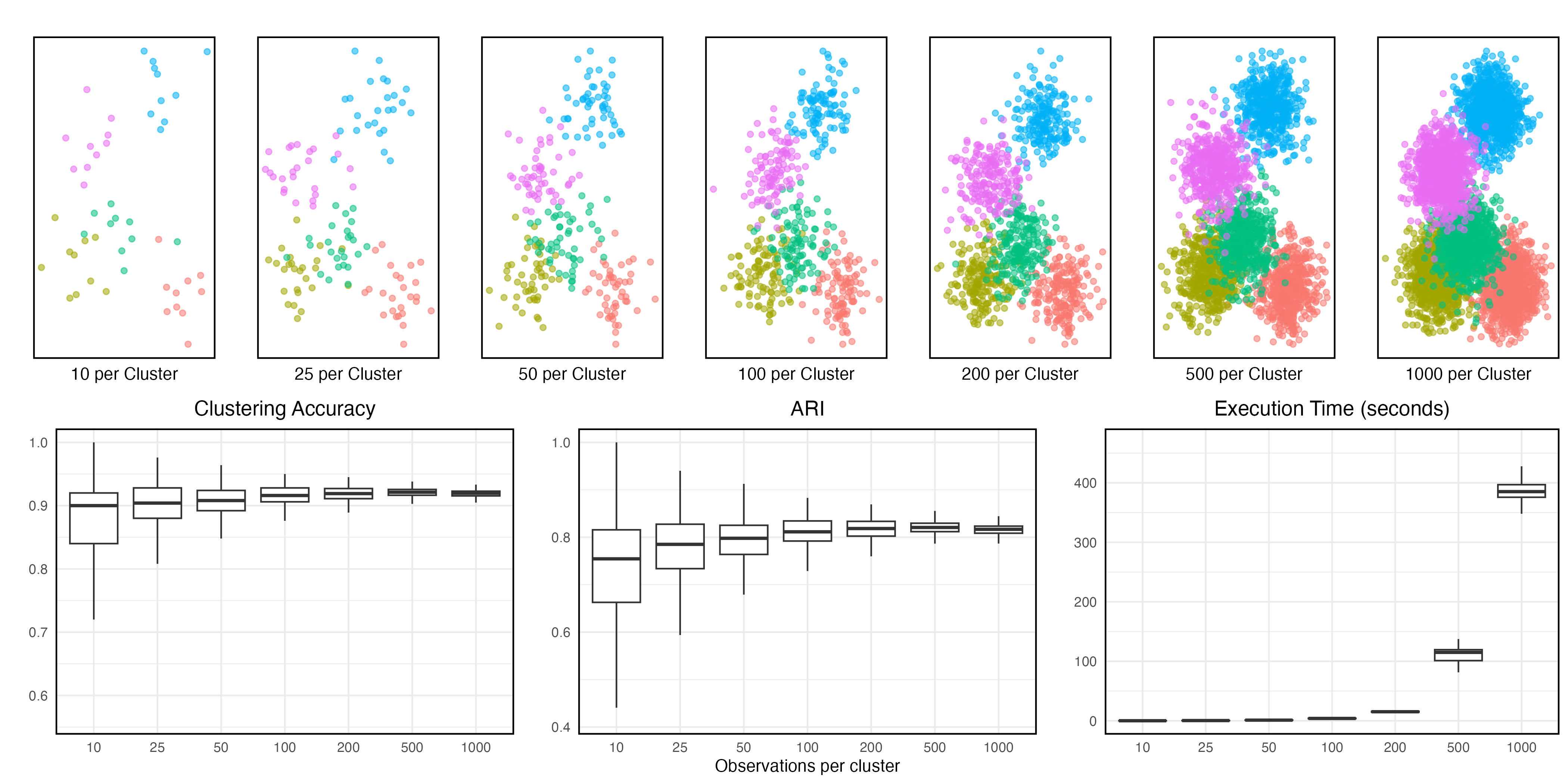}
    \caption{\textcolor{black}{The top row of seven panels are single simulations out of the 500 Monte Carlo Simulations with five cluster centres, each drawn from the numbers of observations from 10 to 1000. The bottom left panel is a boxplot of the clustering accuracies. The bottom centre panel is a boxplot of the ARIs associated with each Monte Carlo simulation at each sample size. The bottom right panel is a boxplot of the execution of time (in seconds) of each Monte Carlo replication at each sample size.}}
    \label{fig:cont_MC}
\end{figure}

\textcolor{black}{Monte Carlo simulations were conducted to discern the performance of KDSUM metric with the modified $k$-means algorithm for distance matrices for varying cluster sizes. This iteration encompasses five clusters generated from a multivariate normal distribution with fixed cluster centres spanning $[0, 10]$. 500 simulations were conducted with sample sizes of 10, 25, 50, 100, 200, 500, and 1000 observations, with single simulation examples shown in the upper panels of Figure \ref{fig:cont_MC}. Fixed cluster centres were maintained for all simulations. We note that our implementation of the KDSUM metric methodology is approximately 10 times slower than typical mixed-type metrics; however, with improved coding, the KDSUM methodology execution time could be improved.\\}

\textcolor{black}{ARI, CA, and execution time were calculated for each iteration at sample sizes of 10 to 1000 observations per cluster, and are shown in Figure \ref{fig:cont_MC}.  We can see that the variability of CA decreases as sample size increases, while the median value increases up to 500 observations. At 1000 observations, we see that the median value decreases due to the increasing number of observations in the overlap of clusters. A similar result in CA can be seen in the ARI results. The execution time increases exponentially, where the median execution time for 1000 observations per cluster simulation is approximately 390 seconds or 6.5 minutes.\\}

\begin{figure}[H]
    \centering
    \includegraphics[width = \textwidth]{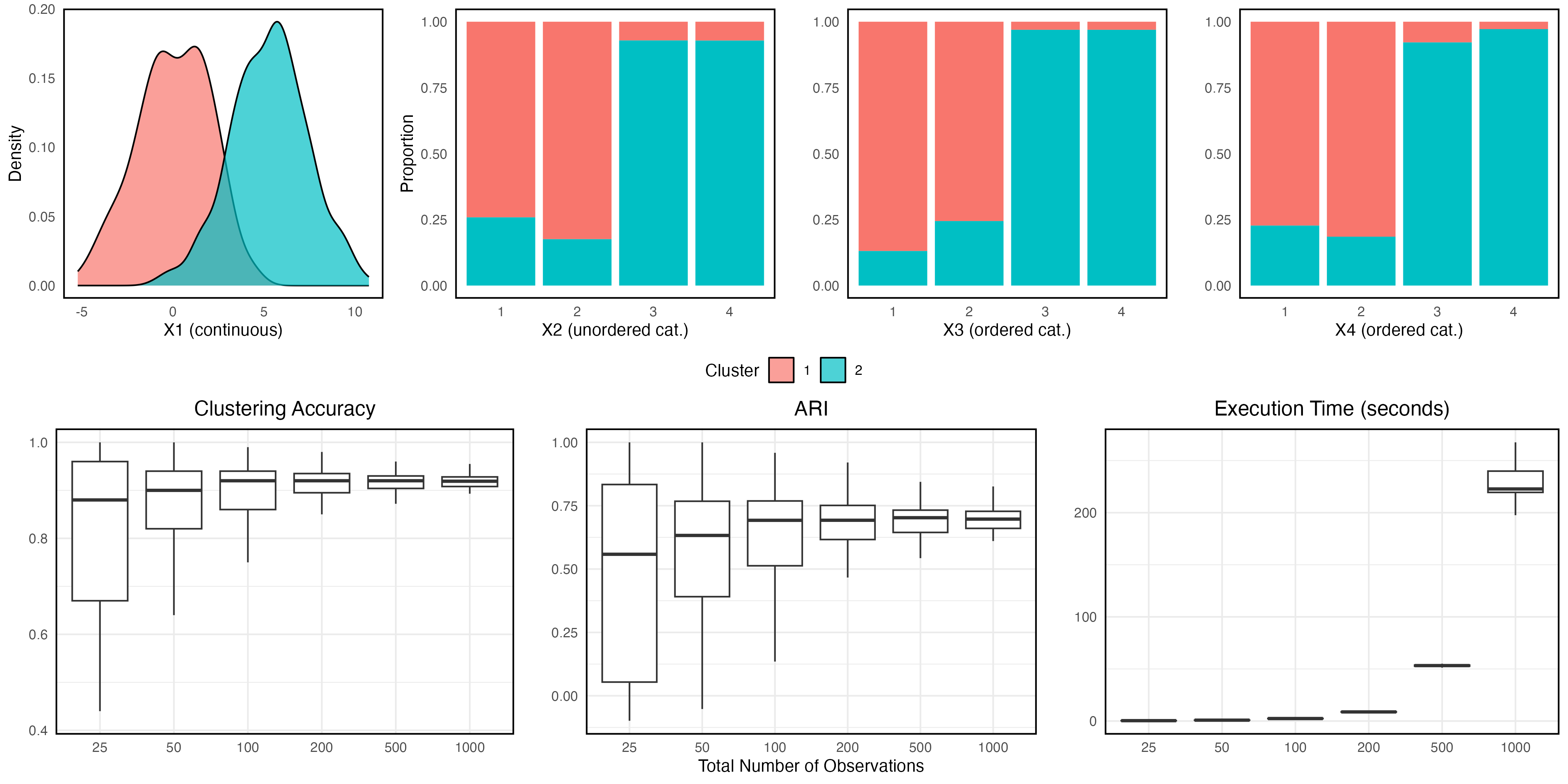}
    \caption{\textcolor{black}{The top row of four panels are the marginal distributions of a single mixed-type data simulation out of the 500 Monte Carlo Simulations. The bottom left panel is a boxplot of the clustering accuracies. The bottom centre panel is a boxplot of the ARIs associated with each Monte Carlo simulation at each sample size. The bottom right panel is a boxplot of the execution of time (in seconds) of each Monte Carlo replication at each sample size.}}
    \label{fig:mix_MC}
\end{figure}

\textcolor{black}{As we demonstrated the performance of this algorithm for varying sample sizes for continuous-only data, we conducted a simulation study on mixed-type simulated data. The continuous variables are drawn from a normal mixture model, whereas categorical variables follow a multinomial mixture model. Overlap between two clusters for continuous variables is the area of the overlapping region defined by their densities, and for categorical variables, the summed height of overlapping segments is defined by their point masses. The overlap for all variables is set to $20\%$, and the two cluster sizes follow a $1:1$ ratio. The sample sizes span 25, 50, 100, 200, 500, and 1000 observations, where 500 Monte Carlo simulations were executed for each cluster size. The empirical marginal distributions of variables for single simulations are shown in the upper panels of Figure \ref{fig:mix_MC}.\\}

\textcolor{black}{The results of clustering the mixed-type data Monte Carlo simulations are shown in the lower panels of Figure \ref{fig:mix_MC}. We see that the variability of CA and ARI decreases as sample size increases, showing that the increase in information improves the accuracy of clustering when using KDSUM. Similarly to the continuous-only data, there is a slight decrease in the median accuracy and ARI as the sample size increases for larger sample sizes caused by hard partitioning for overlapped clusters. Further, the execution time grows exponentially as sample size increases, where the total time for 1000 observations per cluster is approximately 225 seconds or 3.75 minutes. This execution time is smaller than continuous-only, as there are only two clusters for mixed-type data as opposed to the five clusters for the continuous-only data, where continuous-only data had approximately 2.5 times the number of observations.\\}

\section{Real Data \textcolor{black}{Analysis}}\label{sec:real}
This study utilized a diverse range of data, including continuous, categorical, and mixed-type datasets, to evaluate the KDSUM metric for clustering algorithms. Unless otherwise noted, all datasets used in the study are publicly available through the UCI Machine Learning Repository (Dua \& Graff, 2017). \textcolor{black}{For each dataset, we removed any observation vectors containing at least one \textit{NA} value, as many of the clustering methods used (including KDSUM hierarchical) are not designed for missing data.} \\

\textcolor{black}{Continuous-only datasets include the Body dataset with 24 continuous variables and 507 observations with 2 classes, from \texttt{R}-package \texttt{gclus} (Hurley, 2019). Two additional continuous-only datasets are the Wine dataset with 178 observations of 13 continuous variables and three classes, and Iris dataset with  150 observations with four continuous variables and a classification column containing three distinct classes.} \\

\textcolor{black}{Categorical-only datasets include} the Soybean dataset \textcolor{black}{(Soybean L.) was also included, which contains 307 observations and a mix of 35 ordered and unordered categorical variables and 18 classes, as well as its smaller version (Soybean S.) consisting of four classes and 45 observations. The categorical Breast Cancer \textcolor{black}{(Breast)} dataset with nine ordered categorical variables and two classes, and the Congressional Vote dataset (Vote) was also used, with 435 observations and 15 variables, with two classes. Additionally, we use the Zoo dataset with 101 observations consisting of twelve binary variables and one ordered categorical variable.} \\

\textcolor{black}{Mixed-type variable datasets used include} The Australian Credit \textcolor{black}{(Credit)} and Auto MPG dataset \textcolor{black}{(Auto)}, which have a mix of continuous and categorical variables and two classes each. \textcolor{black}{The Credit dataset consists of 690 observations and 14 variables, 8 of which are treated as unordered categorical and 6 as continuous. The Auto dataset consists of 398 observations and 7 variables (after the car name variable was removed), 1 of which is treated as unordered categorical, 1 as ordered categorical, and 5 as continuous.} For the Auto MPG dataset, the predicted class was a continuous variable (miles per gallon), and was partitioned into 2 distinct classes with an approximately even dispersion of observations into the two classes, namely, miles per gallon $<22$ and $\geq22$. 

\subsection{\textcolor{black}{Results}}

\textcolor{black}{We present the results of ten real datasets in Table \ref{tab:results_real}.} The results of the experiments demonstrate the \textcolor{black}{improvements} of the KDSUM metric in terms of CA \textcolor{black}{and ARI}. \textcolor{black}{For the Body dataset, the Gaussian Mixture model outperformed KDSUM by approximately $4.30\%$ in terms of CA and $0.159\%$ in terms of ARI, but KDSUM outperformed the remaining three methods by at least $6.3\%$ in terms of CA and $0.204\%$ for ARI.} For the Wine dataset, \textcolor{black}{KDSUM tied GMMs for highest CA and ARI, and outperformed all other methods by at least $5.1\%$ for CA and $0.139$ for ARI}. \textcolor{black}{For the Iris dataset, KDSUM did comparably well to PAM and $k-$means, but was outperformed by the two other methods, which we discuss after}. \textcolor{black}{For both the Soybean small dataset, all methods performed equally as well, with the exception of PAM with Gower's distance, which did slightly worse. For the large version of the Soybean dataset, KDSUM outperformed all other methods by at least $9.9\%$ in terms of CA and $0.201$ for ARI. For the Zoo dataset, KDSUM outperformed all others by at least $3.0\%$ for CA, and $0.082$ for ARI, while with the Breast dataset, KDSUM performed just slightly poorer than hierarchical clustering with Gower's distance, and outperformed the other three methods by at least $0.2\%$ for CA and $0.011$ for ARI. For the Vote dataset, KDSUM outperformed all other methods by at least $0.4\%$ for CA and $0.019$ for ARI, while for the Auto dataset, KDSUM outpeformed all other methods by at least $0.2\%$ for CA and $0.08$ for ARI. Lastly, for the Credit dataset, KDSUM outperformed all method by at least $2.3\%$ for CA and $0.056$ for ARI. } \\ 
\begin{table*}
\centering
\caption{\textcolor{black}{Classification results on real data. The KDSUM metric with aggolomerative hierarchical clustering was compared to Euclidean or Gower's distance with hierarchical clustering (HC-E / HC-G) and Partitioning around Medoids (PAM-E / PAM-G), $k-$means, $k-$modes, $k-$prototypes ($k-$proto), Gaussian Mixture Model (GMM), ROCK, and clustMD. For hierarchical clustering, the linkage that provides the best results is reported in parentheses.}}
{\color{black}\begin{tabular}{cccc|cccc}
\hline
\hline
Data & Model & CA & ARI & Data & Model & CA & ARI\\
\hline
\hline
Body        & KDSUM (Average)  & 0.935 & 0.756 &    Zoo & KDSUM (Complete) & 0.921 & 0.940 \\
(cont.)& PAM-E & 0.872 & 0.552  &              (cat.) & PAM-G & 0.813 & 0.662 \\
            & $k-$means & 0.864 & 0.529 &           & $k-$modes & 0.800 & 0.647 \\
            & HC-E (Ward) & 0.868 & 0.540 &       & HC-G (Complete) & 0.881 & 0.847 \\
            & GMM & 0.978 & 0.915 &                 & ROCK & 0.891 & 0.858 \\ \hline
Wine        & KDSUM (Ward)  & 0.978 & 0.929      & Breast  & KDSUM (Ward)  & 0.957 & 0.836  \\ 
(cont.)& PAM-E & 0.708 & 0.371                 & (cat.) & PAM-G & 0.955 & 0.825  \\
            & $k-$means & 0.702 & 0.377 &           & $k-$modes & 0.933  & 0.745  \\
            & HC-E (Ward) & 0.927 & 0.790 &     & HC-G (Ward) & 0.968  &0.874 \\
            & GMM & 0.978 & 0.929 &                 & ROCK & 0.686 & 0.041 \\ \hline
Iris       & KDSUM (Ward) & 0.887 & 0.718       &  Vote & KDSUM (Ward) & 0.914 & 0.684 \\
(cont.)           & PAM-E & 0.893 & 0.730      & (cat.) & PAM-G & 0.867 & 0.535 \\
            & $k-$means & 0.893 & 0.730 &           & $k-$modes & 0.867 & 0.535 \\
            & HC-E (Ward)  & 0.907 & 0.759 &        & HC-G (Average) & 0.910 & 0.665 \\
            & GMM & 0.967  & 0.904 &                & ROCK & 0.828 & 0.504 \\ \hline
Soybean S. & KDSUM (Average) & 1.000 & 1.000  & Auto  & KDSUM (Average) & 0.913 & 0.682 \\
 (cat.) & PAM-G & 0.936 & 0.820 & (mix.) & PAM-G & 0.829 & 0.431 \\
            & $k-$modes & 1.000 & 1.000 &       & $k-$Proto & 0.888 & 0.600 \\
            & HC-G (Complete) & 1.000 & 1.000 & & HC-G (Average) & 0.911 & 0.674 \\
            & ROCK & 1.000 & 1.000 & & clustMD & 0.880 & 0.557 \\ \hline
Soybean L.  & KDSUM (Ward) & 0.792 & 0.577 & Credit & KDSUM (Ward) & 0.817 & 0.401 \\
(cat.) & PAM-G & 0.693 & 0.376 & (mix.) & PAM-G & 0.794 & 0.345 \\
            & $k-$modes & 0.673 & 0.320 & & $k-$Proto & 0.793 & 0.342 \\
            & HC-G (Ward) & 0.628 & 0.315 & & HC-G (Ward) & 0.746 & 0.241 \\
            & ROCK & 0.679 & 0.327 & & clustMD & 0.564 & 0.004 \\ 
\hline
\hline
\end{tabular}}
\label{tab:results_real}
\end{table*}
\textcolor{black}{For the Iris dataset,} the KDSUM method did not perform as well as the competing methods. It is worth mentioning that if we only consider one column (petal width), \textcolor{black}{KDSUM (Ward)} achieved CA and ARI of $0.960$ and $0.886$, respectively, which is the highest value obtained from any combination of variables in this dataset. While the obvious cluster of the setosa species class label was correctly identified, the overlapping nature of the remaining two species, setosa and versicolor, led the KDSUM method to incorrectly classify more than the other methods. Improving the effectiveness of the KDSUM metric for handling overlapping clusters is an active area of consideration. 

\section{Conclusion} \label{sec:conclusions}
In this study, we proposed a novel kernel distance metric for effectively handling mixed-type data. Specifically, we developed a metric based on \textcolor{black}{using } kernel function\textcolor{black}{s as similarity functions, where we proved that the Gaussian, Aitken and Wang \& van Ryzin kernels are similarity functions.} To ensure the viability of our KDSUM metric, we rigorously proved that it satisfies all necessary properties of a distance metric, including non-negativity, symmetry, the triangle inequality, and the identity of indiscernibles. In doing so, we established the theoretical foundation for our KDSUM metric \textcolor{black}{as a shrinkage methodology} and demonstrated its potential for accurately capturing the distances between mixed-type data points. \\

We conducted extensive experiments on both simulated and \textcolor{black}{real} data to evaluate the effectiveness of our KDSUM metric compared to existing mixed-type data metrics and state-of-the-art clustering algorithms designed to handle mixed-type data. Using agglomerative hierarchical clustering techniques, we assessed the performance of our KDSUM metric in terms of CA and the ARI. The KDSUM \textcolor{black}{metric in hierarchical clustering} outperformed existing mixed-type data metrics and achieved competitive results compared to state-of-the-art clustering algorithms. Although most existing metrics employ an additive structure for each variable type, which is similar to the KDSUM method, none of the methods analyzed utilize kernels or kernel smoothing techniques to eliminate irrelevant variables for clustering. Instead, they rely on \textcolor{black}{either} parametric approaches that require data transformations through importance weighting of categorical variables that can be controlled by the user directly or estimated using optimization techniques\textcolor{black}{, or nonparametric approaches that fail to adapt to the underlying data generating process for metric calculations or clustering approach}. \\

\textcolor{black}{Early versions of this methodology used maximum cross-validated bandwidths from a mixed-type joint kernel density estimated (Hall, Racine, and Li, 2004). This approach demonstrated good results, but further investigation into replacing the likelihood function with a similarity function demonstrated increased improvements. Thus, the results of using a likelihood optimization approach were not included.\\}

This paper demonstrates the first steps towards a generalized distance for mixed-type data. Some improvements to the methodology are possible. We have calculated \textcolor{black}{similarities and} distances orthogonally to the clustering algorithm. An investigation of \textcolor{black}{a wider range of} bandwidth selection procedures for various distance-based clustering algorithms is also future work. While agglomerative hierarchical clustering was preferred for this study, a new or existing algorithm \textcolor{black}{that incorporates kernels into the clustering loss function} may further enhance the classification and clustering of mixed-type data with a kernel distance metric. A detailed analysis of clustering algorithms that require dissimilarity matrices as input and determining the optimal clustering algorithm that pairs with kernel distance metrics is also future work. Moreover, we identified several promising directions for future research, including applying \textcolor{black}{investigating the effects of using various continuous and categorical kernels on kernel metric calculations, developing numerical methods for determining the optimal number of clusters, and using} kernel metrics \textcolor{black}{for} fuzzy clustering algorithms. By exploring these research directions, we can further explore the applicability and effectiveness of our KDSUM method for clustering mixed-type data. \\




\section*{Data Availability}
The datasets analyzed during the current study are publicly available in the UCI Learning Repository.

\section*{Code Availability}
All code is available upon request from the contact author. The software used for this research is described in Appendix \ref{appendixSoftware}.

\section*{Conflict of Interest}

The authors declare they have no conflict of interest.

\nocite{*}
\printbibliography

\section*{\textcolor{black}{Appendix}}

\appendix

\section{\textcolor{black}{Software and R-packages}}\label{appendixSoftware}

All simulation were ran in $R$-Studio (R Core Team, 2023), using version 4.3.1. 

\textcolor{black}{The mixed-type metrics in Table \ref{tab:distances} were calculated using the \texttt{daisy} function in the package \texttt{cluster} (Maechler et al., 2022) for Gower's distance, and the \texttt{kmed} package (Budiaji, 2022) for the latter three distance metrics.} \\

\textcolor{black}{The package \texttt{FCPS} in \texttt{R} (Christoph \& Stier, 2021) offers a modified version of the traditional $k-$means algorithm (Hartigan \& Wong, 1979) for clustering using a distance matrix, which is also used in conjunction with KDSUM for some experiments. For comparing clustering methods, continuous datasets are compared against hierarchical clustering techniques with the same linkages, using base package \texttt{stats} and function \texttt{hclust} in  (R Core Team, 2023), Partitioning Around Medoids (PAM) using function \texttt{pam} in package \texttt{cluster} (Maechler et al., 2022), $k-$means in base package \texttt{stats}, and Gaussian Mixture models using function \texttt{Mclust} in package \texttt{mclust} (Scrucca et al., 2016). Categorical datasets are compared against hierarchical clustering techniques with the same linkages, PAM with Gower's distance, $k-$modes using function \texttt{kmodes} in package \texttt{klaR} (Weihs et al., 2005), and ROCK using function \texttt{rockCluster} in package \texttt{cba} (Buchta \& Hahsler, 2022). Mixed datasets are compared against hierarchical techniques with the same linkages, PAM with Gower's distance, $k-$prototypes using function \texttt{kproto} in package \texttt{clustMixType} (Szepannek, 2018), and function / package \texttt{clustMD} (McParland \& Gormley, 2017). The $CA$ value is calculated with both base package \texttt{stats} and function \texttt{matchLabels} in the  package \texttt{Thresher} (Coombes, 2019).} \textcolor{black}{The ARI is calculated using in  using the package \texttt{mclust} (Scrucca et al., 2022). For each simulation in Figures \ref{fig:bwgrid_mix} and \ref{fig:mix_MC}, the data are generated using \texttt{genMixedData} from the \texttt{R} package \texttt{kamila} (Foss \& Markatou, 2018), consisting of one continuous, one unordered, and two ordered categorical variables.}

\end{document}